\documentclass[10pt,twocolumn,letterpaper]{article}
\usepackage{wacv}              %

\usepackage{graphicx}
\usepackage{amsmath}
\usepackage{amssymb}
\usepackage{booktabs}

\usepackage[pagebackref,breaklinks,colorlinks]{hyperref}

\usepackage[capitalize]{cleveref}
\crefname{section}{Sec.}{Secs.}
\Crefname{section}{Section}{Sections}
\Crefname{table}{Table}{Tables}
\crefname{table}{Tab.}{Tabs.}

\usepackage{bbm}
\usepackage{amsmath}
\usepackage{cleveref}
\usepackage{multirow}
\usepackage{booktabs}
\usepackage{float}

\renewcommand{\vec}[1]{\ensuremath{\boldsymbol{#1}}}

\usepackage{amssymb}
\usepackage{utfsym}
\newcommand{\crossmark}{\scalebox{0.75}{\usym{2613}}}

\def\wacvPaperID{9} %
\def\confName{WACV}
\def\confYear{2025}

\begin{document}

\title{Semantic Neural Radiance Fields for Multi-Date Satellite Data}

\author{Valentin Wagner, Sebastian Bullinger, Christoph Bodensteiner, and Michael Arens\\
Fraunhofer Institute of Optronics, System Technologies and Image Exploitation\\
{\tt\small \{firstname.lastname\}@iosb.fraunhofer.de}
}
\maketitle

\begin{abstract}
In this work we propose a satellite specific Neural Radiance Fields (NeRF) model capable to obtain a three-dimensional semantic representation (neural semantic field) of the scene. 
The model derives the output from a set of multi-date satellite images with corresponding pixel-wise semantic labels. We demonstrate the robustness of our approach and its capability to improve noisy input labels. We enhance the color prediction by utilizing the semantic information to address temporal image inconsistencies caused by non-stationary categories such as vehicles. 
To facilitate further research in this domain, we present a dataset comprising manually generated labels for popular multi-view satellite images.
Our code and dataset are available at \url{https://github.com/wagnva/semantic-nerf-for-satellite-data}.	
\end{abstract}

\section{Introduction}

As the number of satellites in orbit continues to rise, high-resolution satellite imagery is more accessible than ever. Image-based reconstruction for creating large-scale environmental models has gained popularity due to its cost-effectiveness compared to dedicated systems like LiDAR. There is also an increasing emphasis on assigning semantic information to 3D environments, which enhances the analysis of urban areas and natural resources, leading to better decision-making in fields such as urban planning, environmental monitoring, and disaster management.

Traditional methods have focused on extracting explicit representations like pointclouds or meshes from the satellite images by matching image features \cite{s2p,ames,vissat,ssr}.
Subsequent works \cite{pointnet++,pt3,kpconv} assign the semantic information to the extracted representation as additional step.
\emph{Neural Radiance Fields} (\emph{NeRF}) \cite{nerf} take a different approach to 3D-reconstruction by leveraging \emph{Multi-Layer-Perceptrons} (\emph{MLP}). 
The scene structure is represented with the weights of the network. 
Aggregation of \emph{MLP} outputs along visual rays allows rendering of novel, unseen views.
\emph{NeRF} are capable to learn and reproduce additional modalities beside color using a shared three-dimensional representation \cite{semanticnerf,beyondrgb}.
Handling multi-date satellite image data introduces additional challenges such as domain-specific camera models, variable lighting, moving shadows and transient objects such as vehicles. 
Existing works \cite{snerf,satnerf,eonerf,sundial} present \emph{NeRF} adaptions to tackle these challenges.
As of our best knowledge this is the first work to present a dual modality \emph{NeRF} to the satellite domain.

\begin{figure}[t]
	\centering
	\includegraphics[width=0.451\columnwidth]{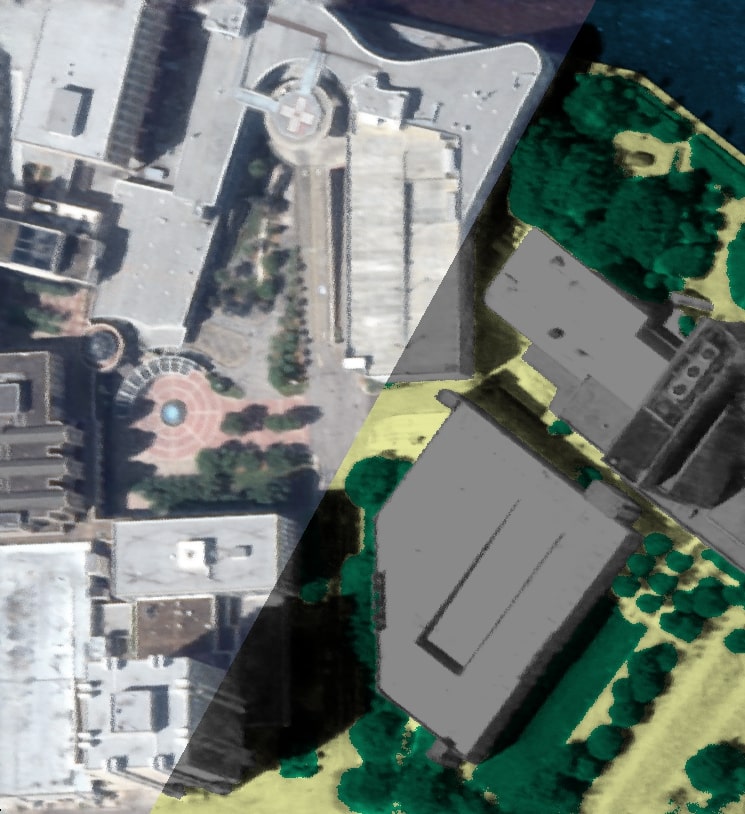}
	\hfil
	\includegraphics[width=0.451\columnwidth]{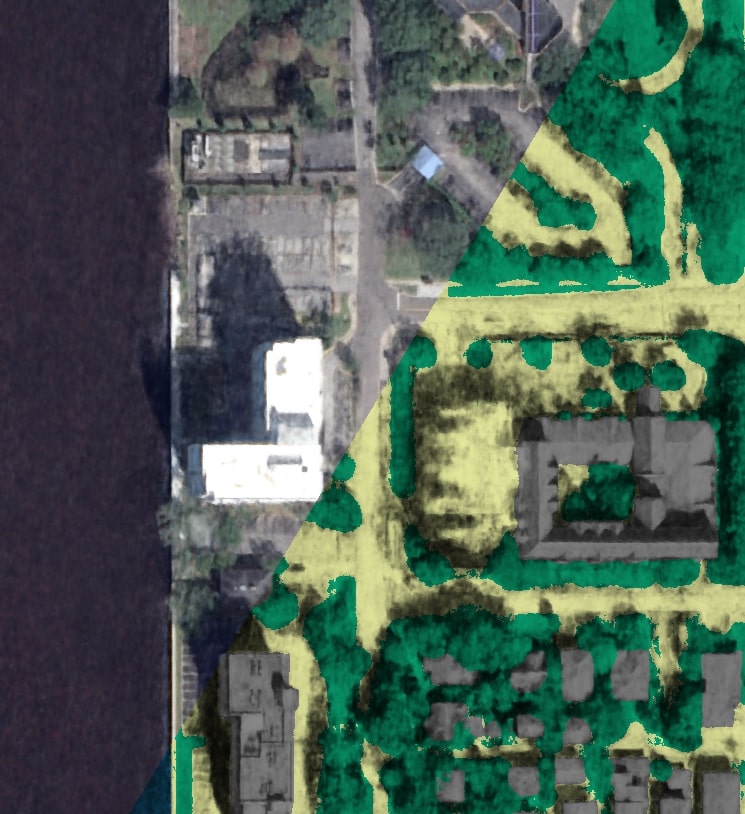}
	\caption{Example output of our proposed satellite-domain-adapted semantic \emph{NeRF} model. Fusion of RGB and semantics allows to render the scene in two distinct modalities for novel, during training unseen views. The semantic visualization is enriched with a three-dimensional structure through combination with a learned lighting component.}
	\label{fig:intro_figure}
\end{figure}

\subsection{Contributions}
Our contributions are as following:~(A) A satellite-domain-adapted semantic \emph{NeRF} capable to fuse semantic and color information onto a unified 3D representation.~(B) Improved color consistency for transient areas such as vehicles in inconsistent images by exploiting semantic information.~(C) Demonstration of the robustness/feasibility of our approach and its capability to improve quality of semantic training data through enforcement of multi-view consistency for the satellite domain.~(D) Publicly released dataset with manually generated pixel-wise annotations for 71 satellite images across four multi-view scenes covering five semantic classes: \emph{Ground}, \emph{Water}, \emph{Vegetation}, \emph{Buildings}, \emph{Vehicles}.~(E) Publicly available code base to reproduce our results and allow further research.

\section{Related Works}

\emph{Semantic-NeRF} \cite{semanticnerf} shows that \emph{NeRF} architectures are capable to simultaneously learn a combined appearance and semantic 3D model for a scene given a set of input color images with corresponding semantic labels.
This allows to render novel, unseen views as color image with a corresponding pixel-wise segmentation image.
The two modalities share the same scene structure, leading to a high cohesion in rendering quality. 
Additionally, they show that the inherent multi-view consistency provided by \emph{NeRF} is able to smooth noisy semantic training labels.

\emph{Shadow-NeRF} \cite{snerf} adapts \emph{NeRF} to the satellite domain by introducing an altitude-based ray sampling and an irradiance lighting model. They separate the static scene color from variable shadows and ambient lighting using the sun direction as input. This allows the network to learn highly variant lighting in its own component. 
\emph{SatNeRF} \cite{satnerf} introduces transient-object filtering and adapts the domain-specific \emph{Rational Polynomial Coefficient} (\emph{RPC}) \cite{rpc} camera model for ray generation, increasing accuracy. 
As satellite datasets are usually multi-date, transient objects are moving inbetween image captures. To reduce artifacts \emph{SatNeRF} \cite{satnerf} introduces an uncertainty term, learned based on an image-specific embedding. 
\emph{EO-NeRF} \cite{eonerf} further iterates on handling the varying lighting conditions. Instead of learning shadows they are dynamically rendered based on the scene geometry and sun direction.
\emph{SUNDIAL} \cite{sundial} models complex illumination to improve dynamic shadow rendering of vegetation and water areas.  
Whereas the results for \emph{EO-NeRF} \cite{eonerf} and \emph{SUNDIAL} \cite{sundial} are promising, no code is publicly available. 
We therefore base our work on \emph{Shadow-NeRF} \cite{snerf} and \emph{SatNeRF} \cite{satnerf}.

\section{Semantic Satellite NeRF}

\begin{figure*}[t!]
	\centering
	\includegraphics[width=\linewidth]{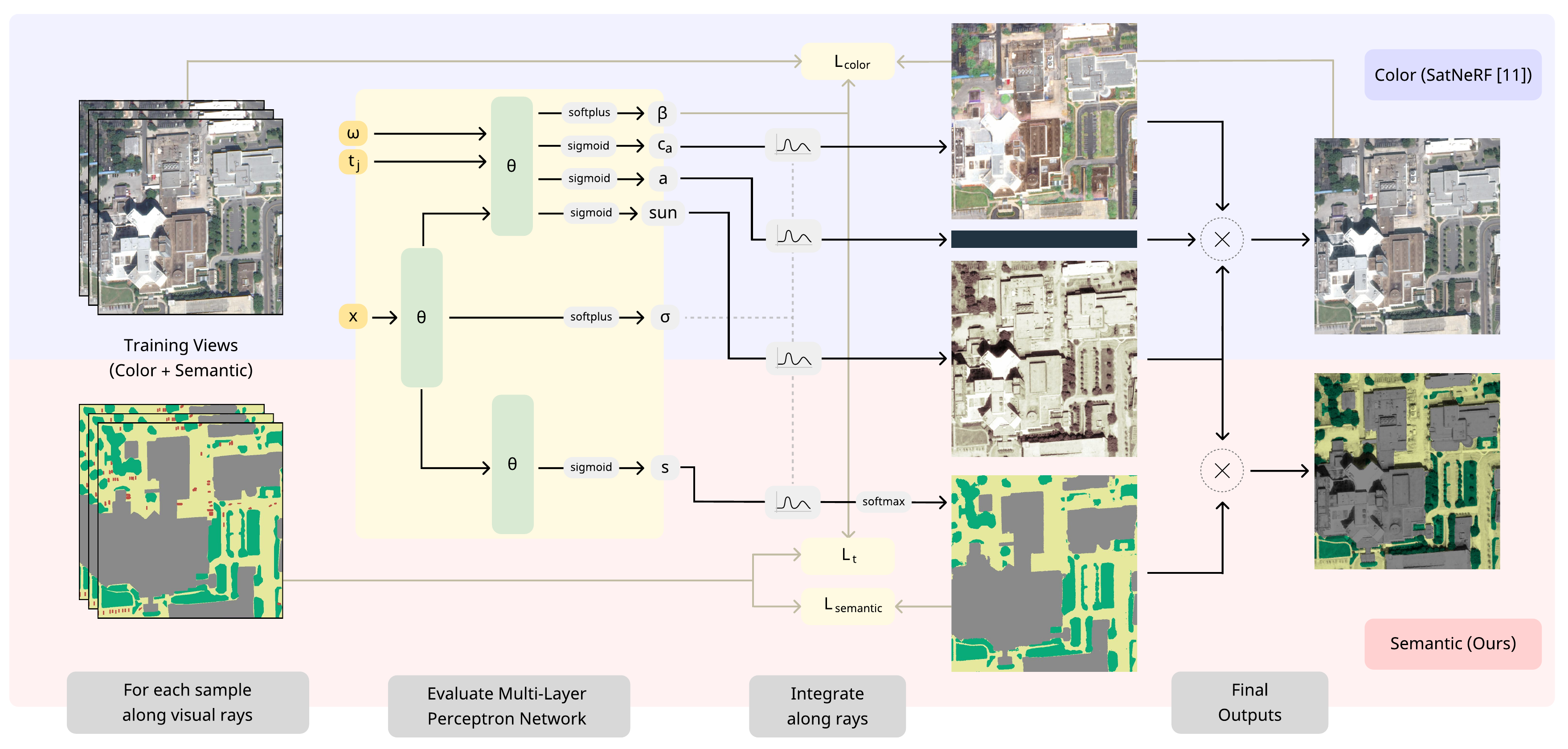}
	\caption{Overview of our proposed model. The satellite-domain-adapted outputs  (\ie elements in the blue area)  are combined using an irradiance lighting model to produce the color rendering as originally proposed by \emph{SatNeRF} \cite{satnerf}. Using an additional semantic head (\ie elements in the red area) our proposed method is able to produce a corresponding semantic pixel-wise labeling. We combine this with the learned lighting scalar to create a three-dimensional semantic visualization. We introduce a transient regularization loss $L_t$ to reduce artifacts in the learned appearance based on the semantic input data. }
	\label{fig:pipeline}
\end{figure*}

An overview of our proposed satellite-domain-adapted semantic \emph{NeRF} can be seen in \cref{fig:pipeline}.
We utilize satellite images with their corresponding semantic pixel-wise annotations to train a dual-modality \emph{NeRF} specifically adapted for satellite-imagery. 
Thus, in contrast to previous approaches, our pipeline renders pixel-wise semantic labels for novel, unseen views of scenes captured by satellite data.

\subsection{Satellite-Domain-Specific Camera Model}
\label{subsec:irradiance_lighting:model}

During the training and testing phase of a \emph{NeRF}, rays are constructed for each pixel of the given images and their corresponding camera poses, traversing a pre-defined, bounded three-dimensional volume. 
As proposed by \emph{SatNeRF} \cite{satnerf} we use the \emph{RPC} \cite{rpc} camera model instead of the commonly used pinhole \cite{pinhole} camera model.
This domain-specific camera model is an approximation of the physical satellite camera. 
We extract it from the satellite image metadata as it is typically used for georegistration of the image. 
To improve accuracy, \emph{SatNeRF} \cite{satnerf} performs \emph{Bundle Adjustment} on the \emph{RPC} models. This process optimizes relative camera poses to increase ray accuracy.

\subsection{Learning Semantic in 3D for Satellite Data}
\emph{NeRF} \cite{nerf} represent a static three-dimensional scene as a continuous volumetric function $\mathcal{F}_{NeRF}$ encoded with a \emph{MLP} network. 
\begin{equation}
	\mathcal{F}_{NeRF}: (\vec{x}, \vec{d}) \mapsto (\sigma, \vec{c})
\end{equation}
Given a 3D point $\vec{x} = (x, y, z)$ and a viewing direction $\vec{d} = (d_x, d_y, d_z)$, the network predicts a density scalar $\sigma$ and a color $\vec{c}=(r, g, b)$.
The scene structure is represented by the density $\sigma$ and used to weigh the influence of the color $\vec{c}$.
\emph{SatNeRF} \cite{satnerf} proposes an expanded satellite-domain-specific architecture. %
\begin{equation}
	\mathcal{F}_{SatNeRF}: (\vec{x}, \vec{\omega}, \vec{t}_j) \mapsto (\sigma, \vec{c}_a, \mathit{sun}, \vec{a}, \beta)
\end{equation}
The scene color $\vec{c}$ is split up into three separate lighting components: the albedo color $\vec{c}_a$, the sun shading scalar $\mathit{sun}$ and the ambient color $\vec{a}$. To predict the variable lighting components $\mathit{sun}$ and $\vec{a}$, the sun direction $\vec{\omega}$ is used as auxiliary input.
The uncertainty $\beta$ is used to reduce artifacts caused by transient objects such as vehicles. It is learned with the per-image embedding $\vec{t}_j$ as auxilliary input. 
Our proposed semantic satellite-adapted network represents the following function $\mathcal{F}_{SemSatNeRF}$.
\begin{equation}
		\mathcal{F}_{SemSatNeRF}: (\vec{x}, \vec{\omega}, \vec{t}_j) \mapsto (\sigma, \vec{c}_a, \mathit{sun}, \vec{a}, \beta, \vec{s})
\end{equation} 
We extend the satellite-domain-adapted architecture from \emph{SatNeRF} \cite{satnerf} with the additional semantic output $\vec{s}$.
Similiar to \emph{Semantic-NeRF} \cite{semanticnerf} the semantic output $\vec{s}$
is represented as a view-invariant function $\mathcal{F}_s(\vec{x}) \mapsto \vec{s}$ of the world coordinates $\vec{x}$ and describes a distribution over $C$ semantic labels as pre-softmax logits.
In addition, we introduce a $\mathit{sigmoid}$ activation function with the intention to increase the generalization across views, using the 
$\mathit{sigmoid}$ activation function as form of normalization. This reduces the ability of single samples to dominate the semantic prediction. 
\begin{equation}
	\vec{s}(\vec{x}) = \mathit{sigmoid}(\mathcal{F}_s(\vec{x}))
\end{equation}
To produce the final result for a specific pixel position, \emph{NeRF}-based pipelines aggregate the outputs of the \emph{MLP} for individual points $\vec{x}$ along a ray $\vec{r}(t) = \vec{o} + t \cdot \vec{d}$ representing the corresponding line of sight. Here, $\vec{o}$ and $\vec{d}$ denote the origin and direction vectors.
\begin{equation}
	\vec{s}(\vec{r}) = \sum_{i=1}^{N} T_i \alpha_i \vec{s}(\vec{x}_i)	
	\label{eq:semantic_rendering}
\end{equation}
The influence of each sample $\vec{x}_i$ along a ray is based on its opacity $\alpha_i$ and transmittance $T_i$.
\begin{equation}
	\alpha_i = 1 - \exp(-\sigma_i\delta_i) \text{ and } T_i = \prod_{j=1}^{i-1} (1 - \alpha_j)
\end{equation}
The distance $\delta_i = t_i -  t_{i-1}$ describes the length traveled along the ray between the previous and current sampled position $\vec{x}$.
Using softmax normalization the combined logits $\vec{s}(\vec{r})$ are converted into probabilities $\hat{\vec{p}}(\vec{r})$. The class with the highest probability is chosen to represent the semantic class of the ray.

\subsection{Semantic Structural Visualization}
\label{subsec:3d_viz}

To visualize the semantic labels for a given image, usually a simple color mapping operation $f_c: \vec{s} \mapsto \vec{RGB}$ is used. 
This visualization is sub-optimal for large areas with the same class, as the structure of the scene is lost. 
For large structures consisting of multiple buildings with large height differences this is especially noticeable, such as in the scenes JAX\_068 and JAX\_214 in \cref{fig:results_labels}.

To handle variance in lighting caused by different sun positions in training images, we use a satellite-domain-adapted irradiance lighting model as proposed by \emph{Shadow-NeRF} \cite{snerf}.
The sample color $\vec{c}(\vec{x}, \vec{\omega})$ is split up into three components: 
scene color $\vec{c_a}(\vec{x})$, lighting scalar $\mathit{sun}(\vec{x}, \vec{\omega})$ and ambient sky color $\vec{a}(\vec{\omega})$.
\begin{equation}
	\vec{c}(\vec{x}, \vec{\omega}) = \vec{c}_a(\vec{x}) \cdot (\mathit{sun}(\vec{x}, \vec{\omega}) + (1 - \mathit{sun}(\vec{x}, \vec{\omega})) \cdot \vec{a}(\vec{\omega}))
	\label{eq:satnerf_lighting_model}
\end{equation}

By referencing the visualization of the $sun(\vec{x}, \vec{\omega})$ lighting component in \cref{fig:results_shadows} we can see that it resembles a grayscale relief map of the scene structure.
We use this to provide a three-dimensional depth to the semantic output. We aggregate the $\mathit{sun}(\vec{x}, \vec{\omega})$ component along rays $\vec{r}$ analog to \cref{eq:semantic_rendering} and combine it with the mapped semantic color $f_c(\vec{s}(\vec{r}))$.
\begin{equation}
	\vec{c}_{s}(\vec{r}) = f_c(\vec{s}(\vec{r})) * \sum_{i=1}^{N} T_i \alpha_i \mathit{sun}(\vec{x}_i, \vec{\omega})
	\label{eq:semantic_viz_shading}
\end{equation}
The resulting visualization of the structural semantic prediction is shown in \cref{fig:results_semantic_shaded}. The commonly flat color visualization is enriched with a three-dimensional structure. Finer details are made visible and objects such as the distinct main building can now be easily identified.

\subsection{Handling Transient Objects}
\label{subsec:handling_transients}
A major issue by processing multi-date satellite images are temporal inconsistencies. Besides changes in the scene geometry, transient objects pose a substantial challenge for the reconstruction process. As transient objects change positions across training images, the network receives contradictory information. 
Since we are interested in reconstructing a consistent model reflecting the stationary scene geometry, transient objects represent a type of noise. 
To improve the model consistency, we use available transient category information contained in the semantic labels to reduce the impact of such objects on the semantic as well as the appearance prediction. 
We separately consider the problem for both modalities as semantic and appearance are learned with their own distinct loss function.

\subsubsection{Handling Transient Objects in RGB Data}
\label{subsubsec:transients_rgb}

\emph{NeRF} are optimized by minimizing the $L2$-Loss between the predicted ray color $\vec{c}(\vec{r})$ and the ground truth pixel color $\vec{c}_{GT}(\vec{r})$.
\begin{equation}
	L_{2}(\mathcal{R}) = \sum_{\vec{r} \in \mathcal{R}}|| \vec{c}(\vec{r}) - \vec{c}_{GT}(\vec{r}) ||_2^2
	\label{eq:rgb_loss_normal}
\end{equation}

To reduce the effect of transient objects in the color prediction \textit{SatNeRF} \cite{satnerf} introduces an uncertainty mechanism. Based on a per-image learned embedding $t_j$ the network predicts an uncertainty factor $\beta$ for each location $\vec{x}$.
The color-loss is decreased depending on the uncertainty, giving the network the ability to selectively reduce the loss of specific rays. 
\begin{equation}
	L_{color}(\mathcal{R}) = \sum_{\vec{r} \in \mathcal{R}} \frac{|| \vec{c}(\vec{r}) - \vec{c}_{GT}(\vec{r}) ||^2_2}{2\beta'(\vec{r})^2} + \left( \frac{\log{\beta'(\vec{r}) + \eta}}{2} \right)
	\label{eq:loss_rgb}
\end{equation}
The second term forces the network to use the uncertainty sparingly preventing the naive solution of $\beta$ converging towards infinity. To prevent negative values in the logarithm we set $\beta'(\vec{r}) = \beta(\vec{r}) + 0.05$ and $\eta = 3$ analog to \emph{SatNeRF} \cite{satnerf}.

The uncertainty $\beta$ gives the network an mechanism to reduce the impact of local outliers in the color training images. Although many transient objects such as vehicles move inbetween images, locations such as parking lots contain vehicles in majority of the images. To achieve our goal of learning a static, transient free representation of the scene we introduce a transient regularization loss. Instead of a purely statistical approach we propose guiding the uncertainty towards transient objects using existing information from the semantical ground truth data.
To this end, we introduce a transient regularization loss as seen in \cref{eq:car_reg_loss}. For rays $\vec{r} \in \mathcal{R}^t$ that are associated with transient ground truth labels, the following loss guides the uncertainty $\beta$ towards $1$. 
\begin{equation}
		L_{t}(\mathcal{R}^t) = \sum_{\vec{r} \in \mathcal{R}^t} || 1.0 - \beta(\vec{r})  ||^2_2 
		\label{eq:car_reg_loss}
\end{equation}
We do not make any assumptions about rays outside of the transient classes, allowing the network to still learn its own uncertainty values. This ensures that the network still handles other occurring phenomena such as chromatic aberration. 
Whereas the uncertainty $\beta$ is theoretically unbounded, the second term of the color-loss in \cref{eq:loss_rgb} punishes high values. The uncertainty scalar $\beta$ naturally resides in the interval from $0$ to around $1$. We therefore guide the network towards $1$ for rays $\vec{r} \in \mathcal{R}^{t}$.

\emph{SatNeRF} \cite{satnerf} proposes using the non-modified color-loss as seen in \cref{eq:rgb_loss_normal} for the first two epochs. This gives the network time to learn an initial understanding of the scene geometry and color. We enable our proposed transient regularization loss after three epochs. 
This gives the network one additional epoch to initialize the uncertainty $\beta$.

\subsubsection{Handling Transient Objects in Semantic Data}
\label{subsubsec:transients_semantic}

Common \emph{NeRF} methods only derive a 3d appearance model of the scene. By introducing a semantic representation, the network now has a secondary goal, requiring an additional loss term.
\emph{Semantic-NeRF} \cite{semanticnerf} uses a cross-entropy-loss $\sum_{c=1}^{C} p^c(\vec{r}) \log{\hat{p}^c(\vec{r})}$ comparing the ground truth label $p^c$ with the predicted semantic probability $\hat{p}^c$ for each class $c$ across all rays $\vec{r} \in \mathcal{R}$.
We define the semantic loss to only include rays from static classes $\vec{r} \in \mathcal{R} \setminus \mathcal{R}^t$. $\mathcal{R}^t$ hereby describes all rays $\vec{r}$ with a semantic class defined as transient. 
\begin{equation}
	L_{semantic}(\mathcal{R} \setminus \mathcal{R}^t) = - \sum_{\vec{r} \in (\mathcal{R} \setminus \mathcal{R}^t )} \sum_{c=1}^{C} p^c(\vec{r}) \log{\hat{p}^c(\vec{r})}
	\label{eq:loss_semantic_modified}
\end{equation}
This effectivelly equals in ignoring parts of the semantic training labels, leaving holes in the pixel-wise label map. The network does not receive any learning guidance for these positions and has to rely on information given by different training views to fill in the missing information. This is made possible by the inherent multi-view consistency mechanism provided by \emph{NeRFs}.

\section{Semantic Multi-View Satellite Dataset}

\begin{figure*}[ht]
	\centering
	\begin{minipage}{0.015\linewidth}
		\centering
		\rotatebox[origin=center]{90}{JAX\_004}
	\end{minipage}
	\begin{minipage}{0.98\linewidth}
		\centering
		\includegraphics[width=.19\linewidth]{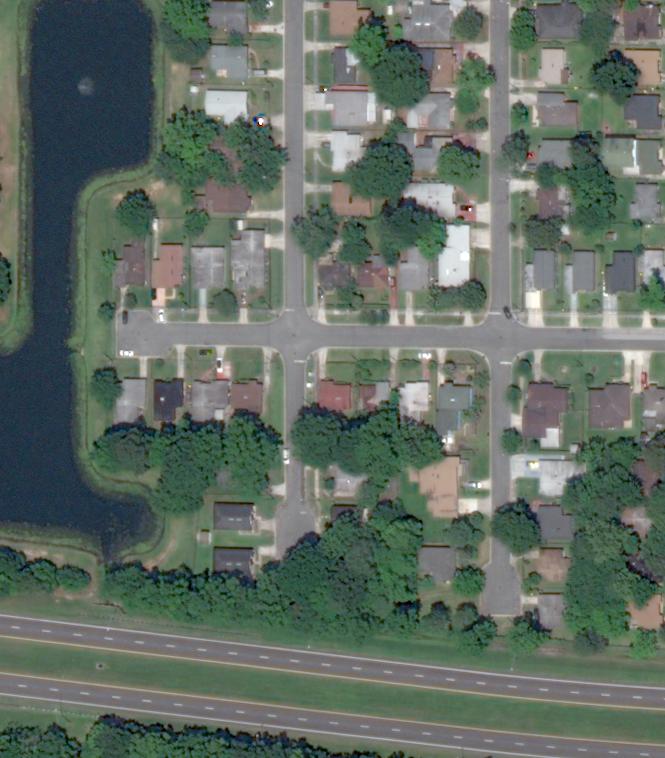}
		\hfill
		\includegraphics[width=.19\linewidth]{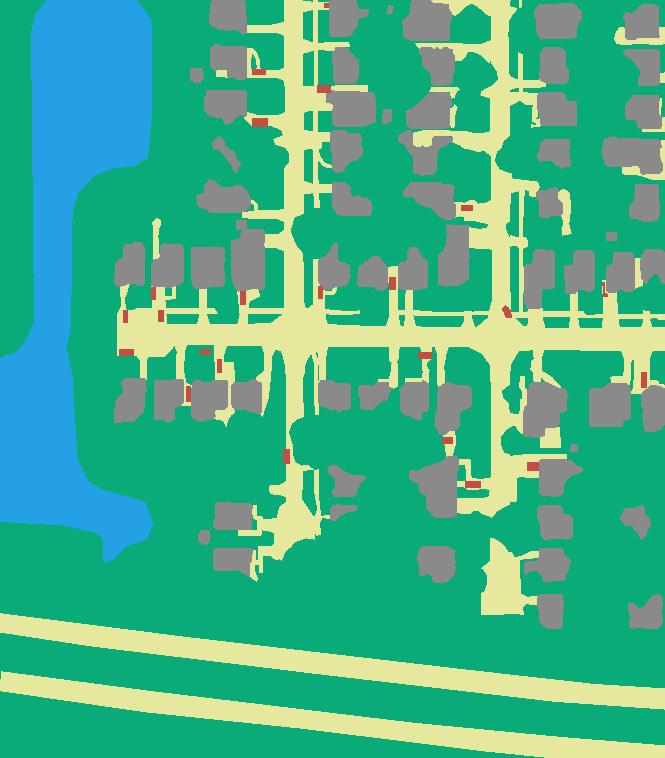}
		\hfill
		\includegraphics[width=.19\linewidth]{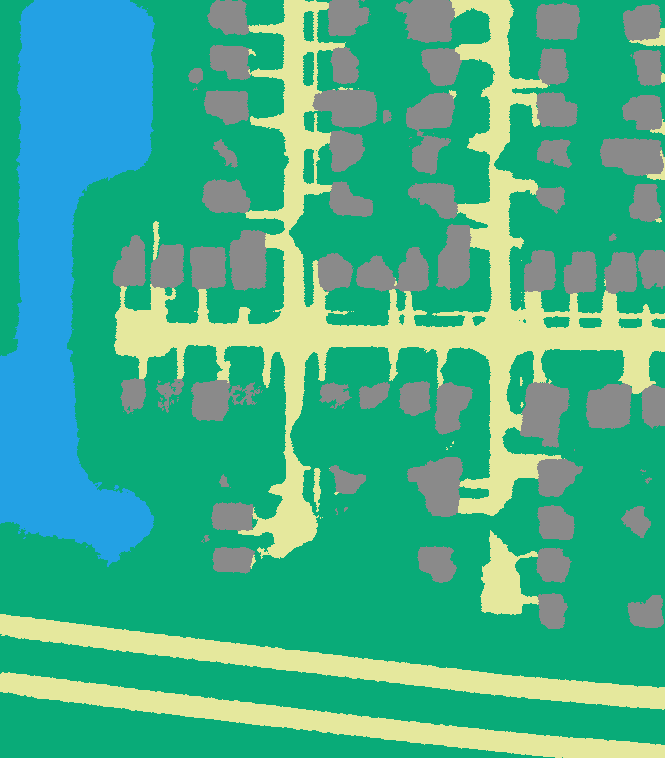}
		\hfill
		\includegraphics[width=.19\linewidth]{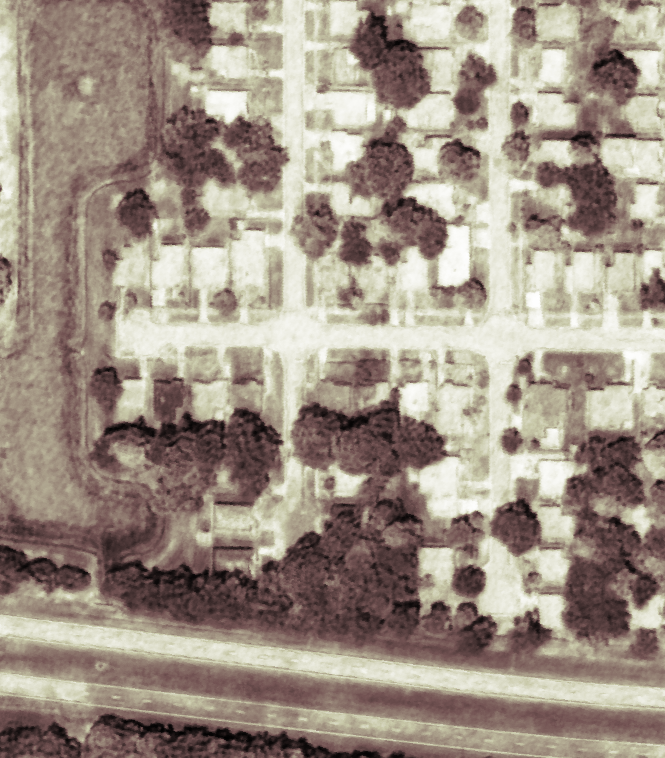}
		\hfill
		\includegraphics[width=.19\linewidth]{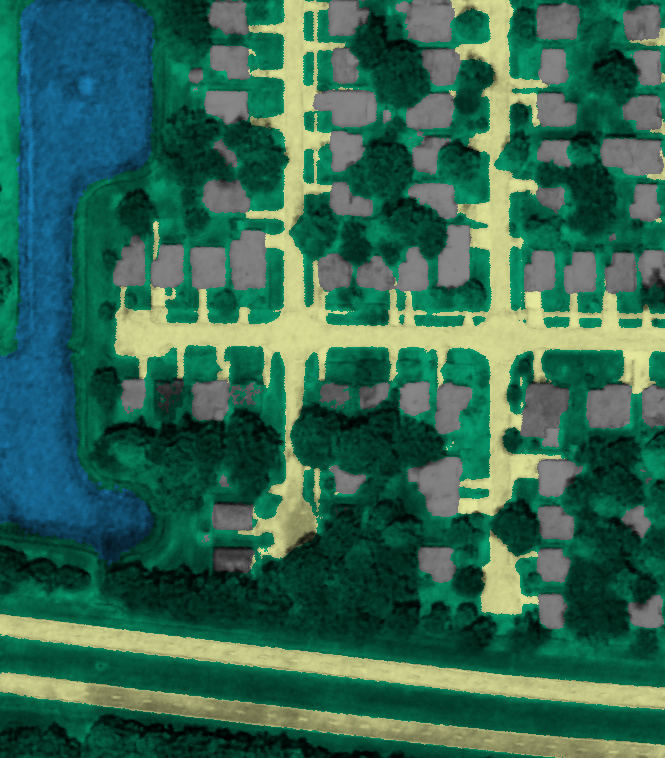}\\
	\end{minipage}\\
	
	\vspace{0.2 cm}
	\begin{minipage}{0.015\linewidth}
		\centering
		\rotatebox[origin=center]{90}{JAX\_068}
	\end{minipage}
	\begin{minipage}{0.98\linewidth}
		\centering
		\includegraphics[width=.19\linewidth]{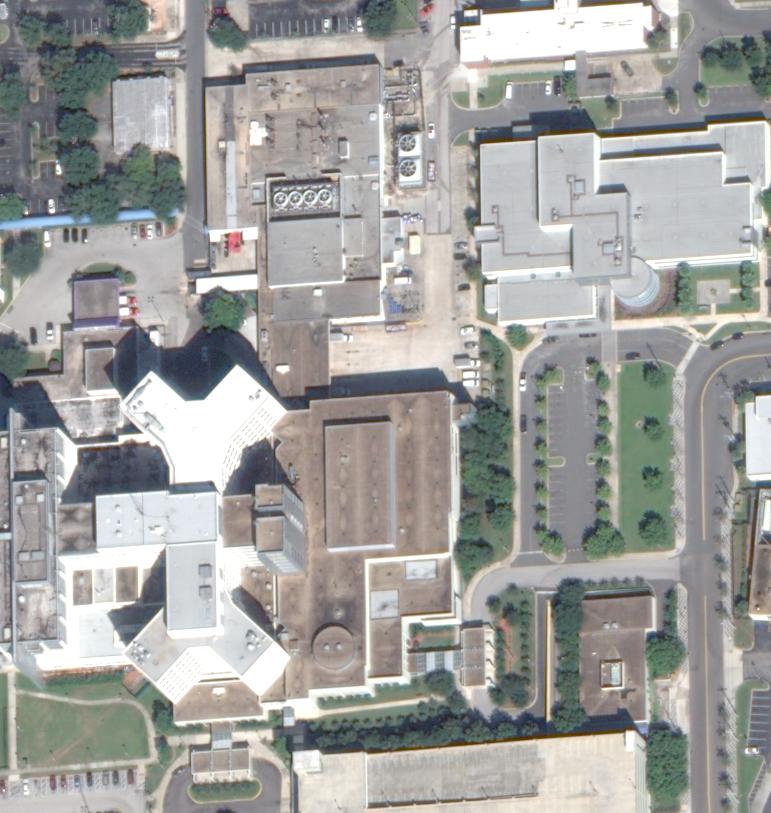}
		\hfill
		\includegraphics[width=.19\linewidth]{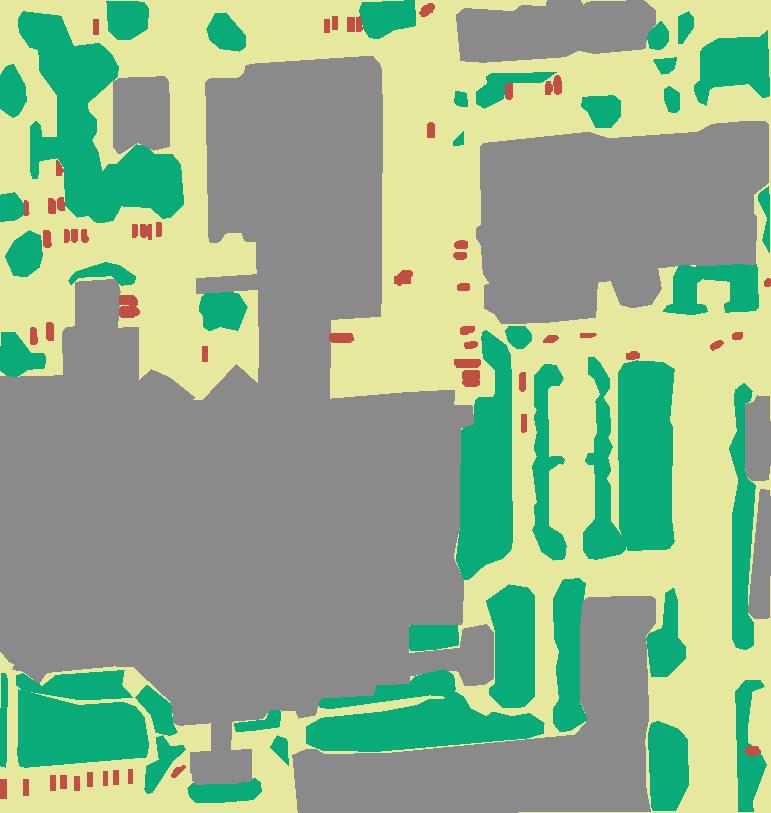}
		\hfill
		\includegraphics[width=.19\linewidth]{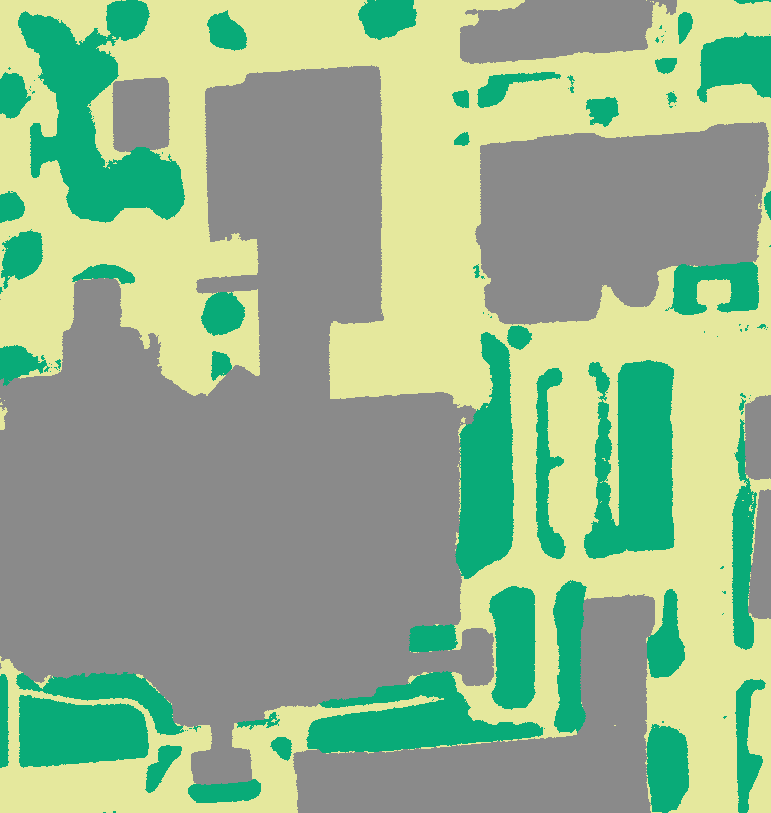}
		\hfill
		\includegraphics[width=.19\linewidth]{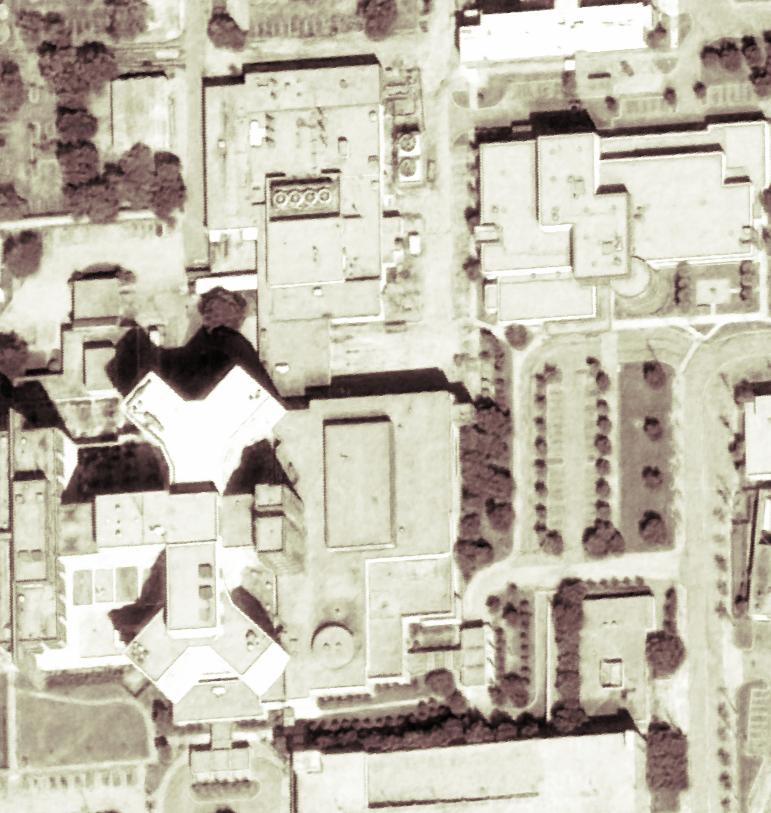}
		\hfill
		\includegraphics[width=.19\linewidth]{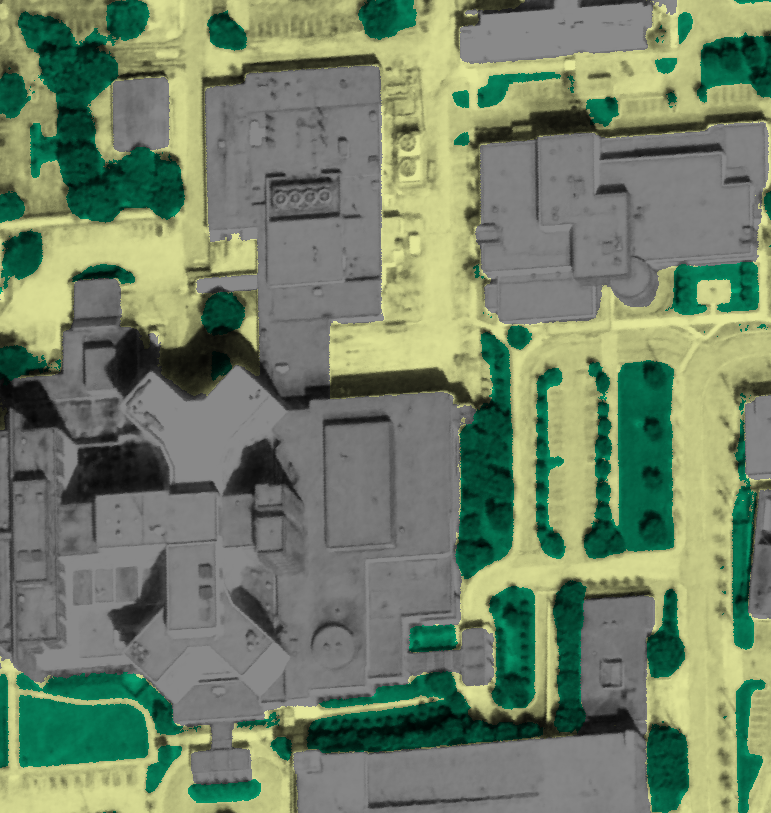}\\
	\end{minipage}\\

	\vspace{0.2 cm}
	\begin{minipage}{0.015\linewidth}
		\centering
		\rotatebox[origin=center]{90}{JAX\_214}
	\end{minipage}
	\begin{minipage}{0.98\linewidth}
		\centering
		\includegraphics[width=.19\linewidth]{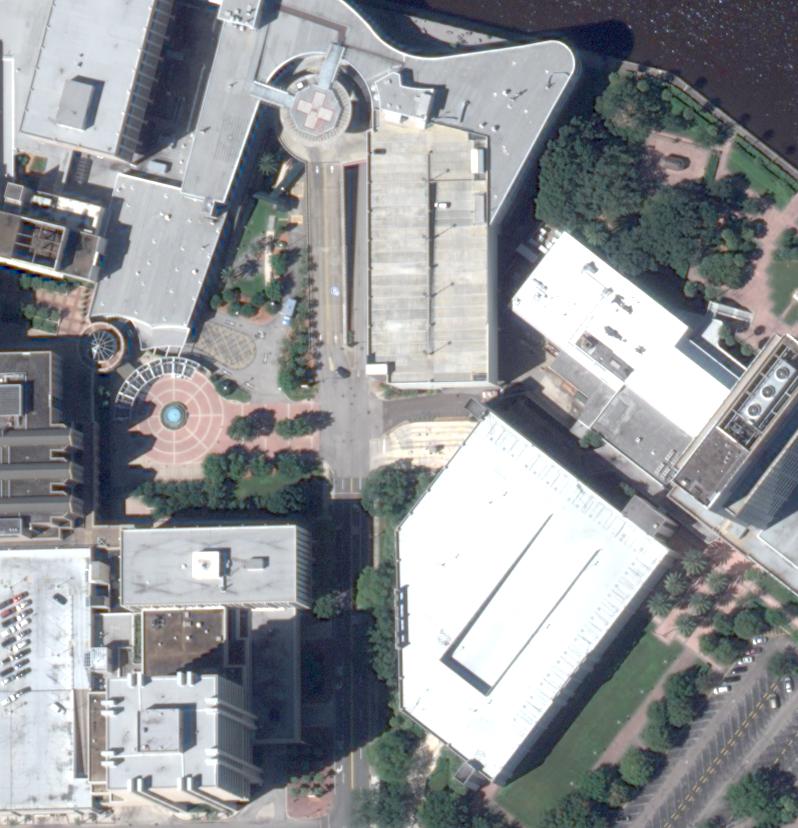}
		\hfill
		\includegraphics[width=.19\linewidth]{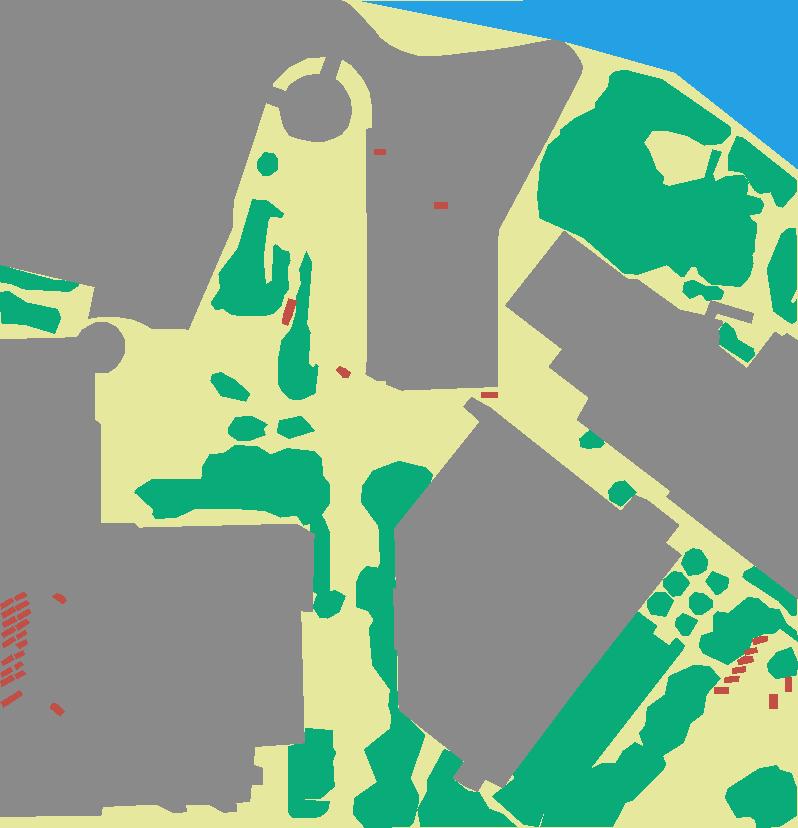}
		\hfill
		\includegraphics[width=.19\linewidth]{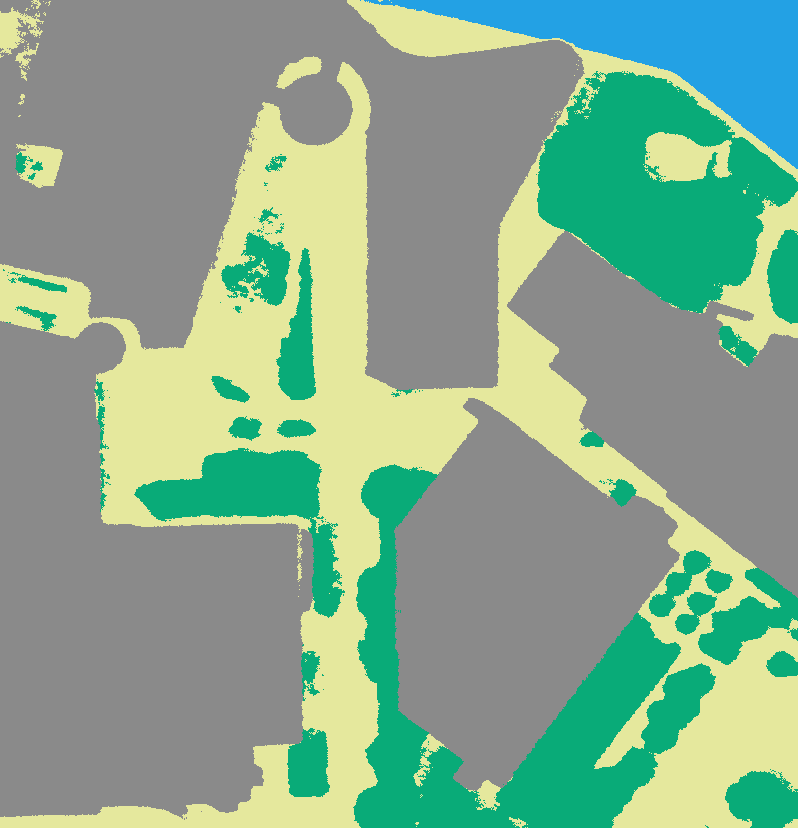}
		\hfill
		\includegraphics[width=.19\linewidth]{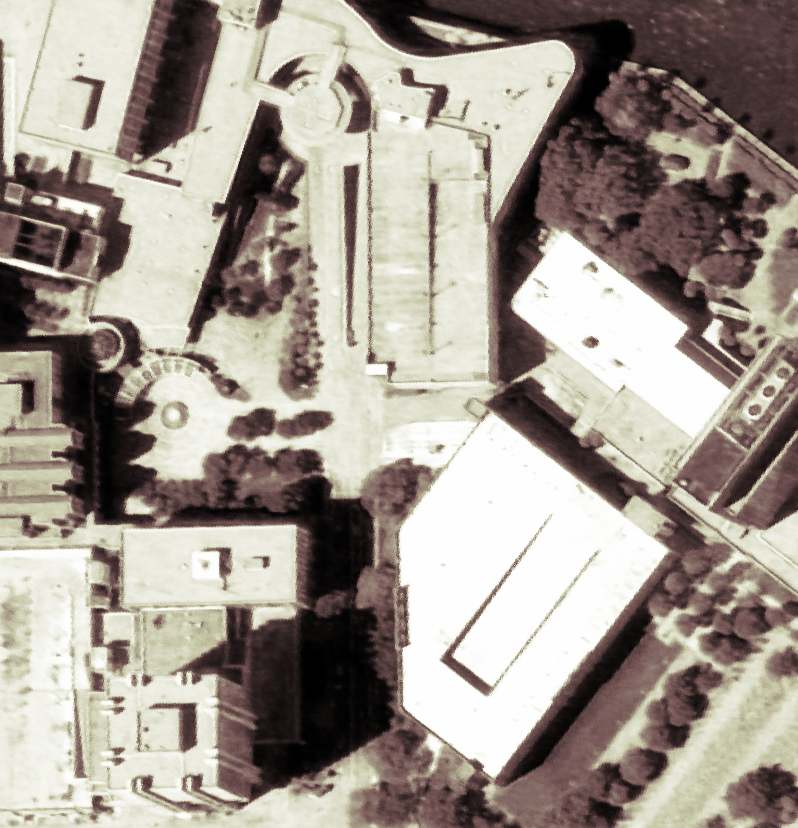}
		\hfill
		\includegraphics[width=.19\linewidth]{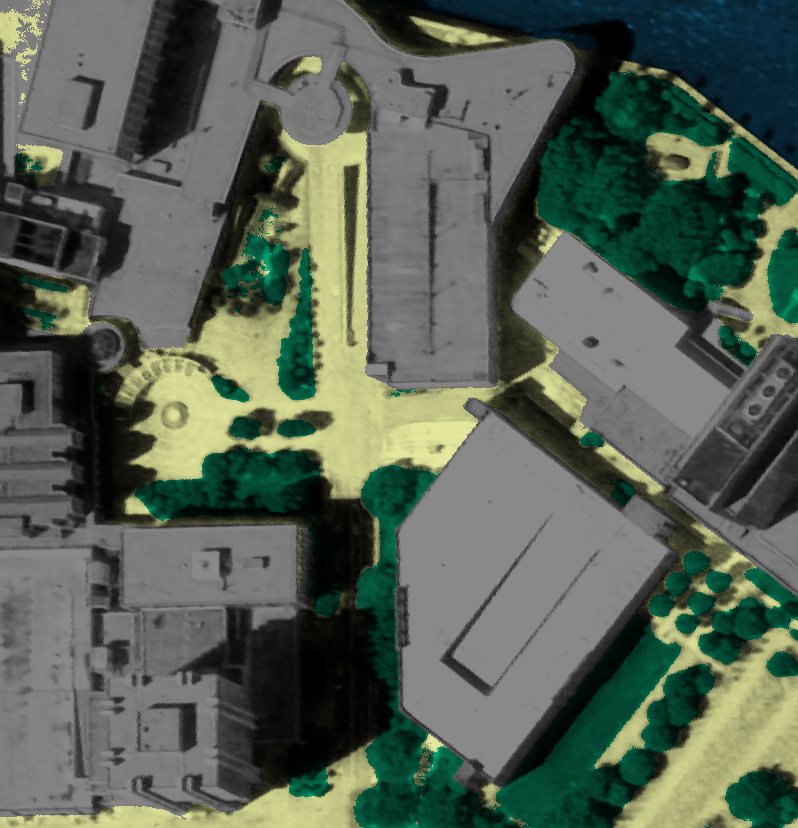}\\
	\end{minipage}\\
	
	\vspace{0.2 cm}
	\begin{minipage}{0.015\linewidth}
		\centering
		\rotatebox[origin=center]{90}{JAX\_260}
	\end{minipage}
	\begin{minipage}{0.98\linewidth}
		\centering
		\subcaptionbox{RGB Ground Truth.\label{fig:results_rgb}}{\includegraphics[width=.19\linewidth]{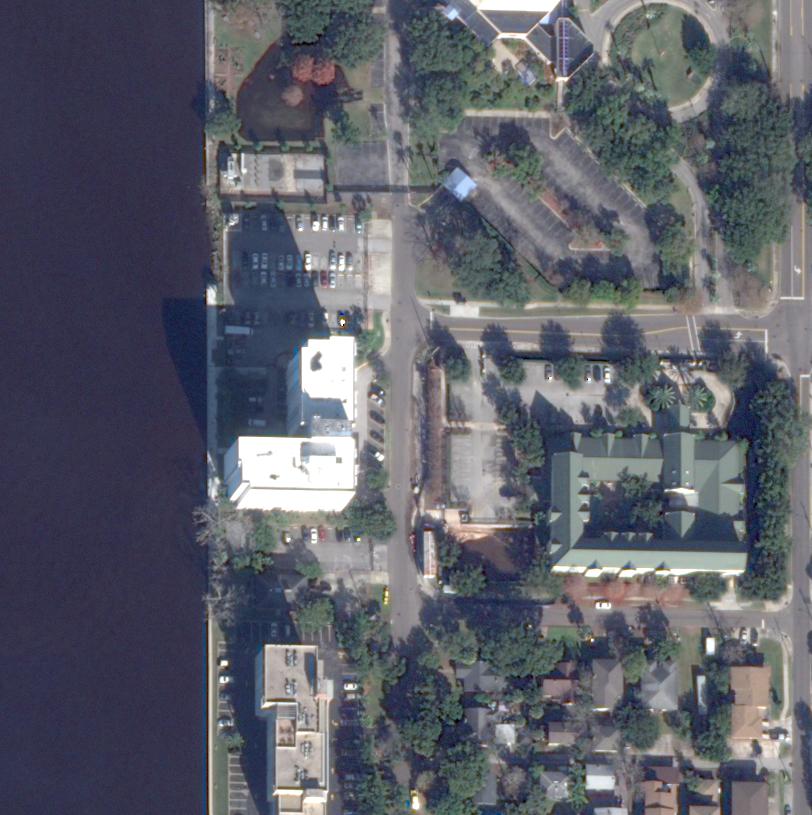}}%
		\hfill
		\subcaptionbox{Our Annotations.\label{fig:results_labels}}{\includegraphics[width=.19\linewidth]{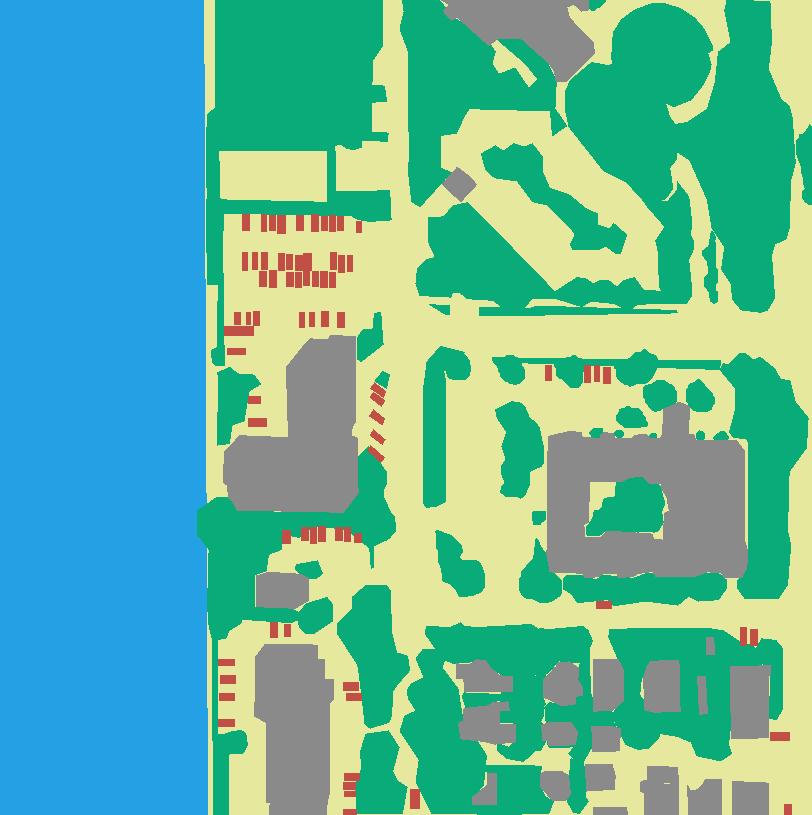}}%
		\hfill
		\subcaptionbox{Predicted Semantic.\label{fig:results_semantic}}{\includegraphics[width=.19\linewidth]{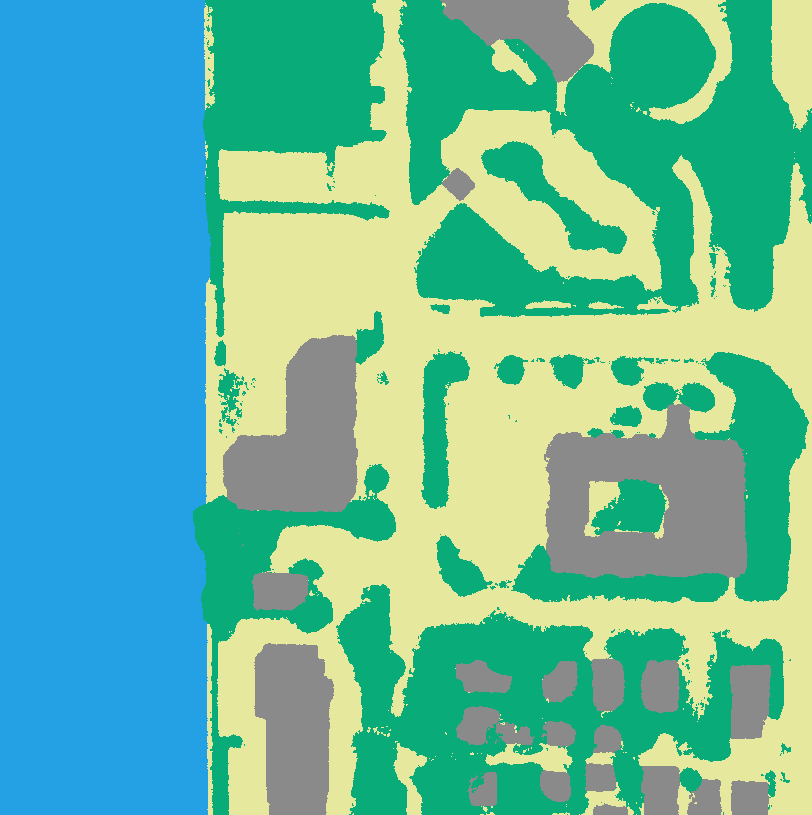}}%
		\hfill
		\subcaptionbox{Predicted Lighting.\label{fig:results_shadows}}{\includegraphics[width=.19\linewidth]{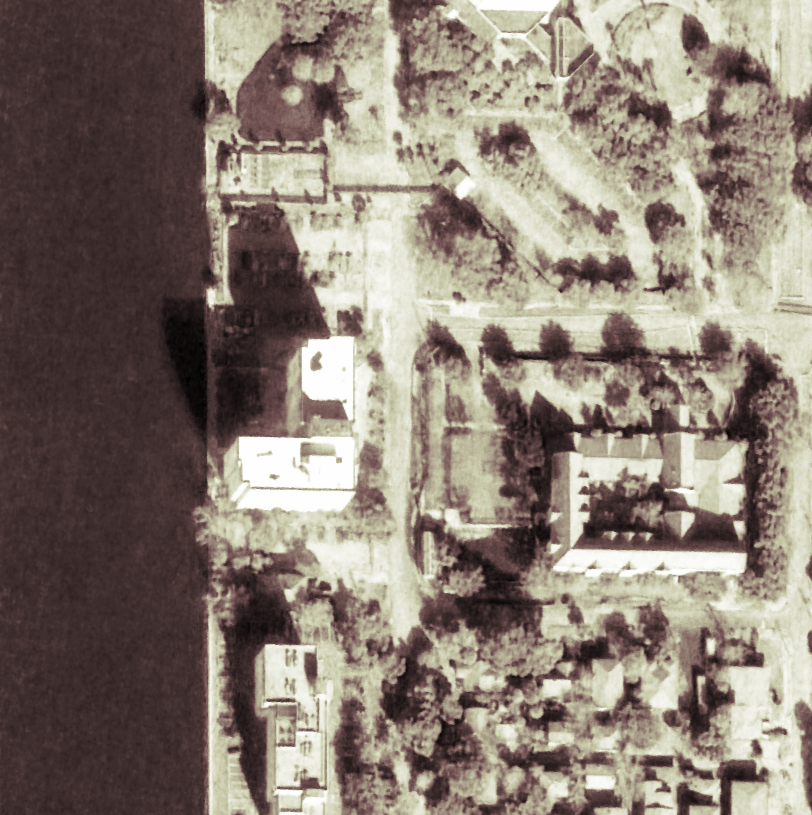}}%
		\hfill
		\subcaptionbox{Semantic Visualization.\label{fig:results_semantic_shaded}}{\includegraphics[width=.19\linewidth]{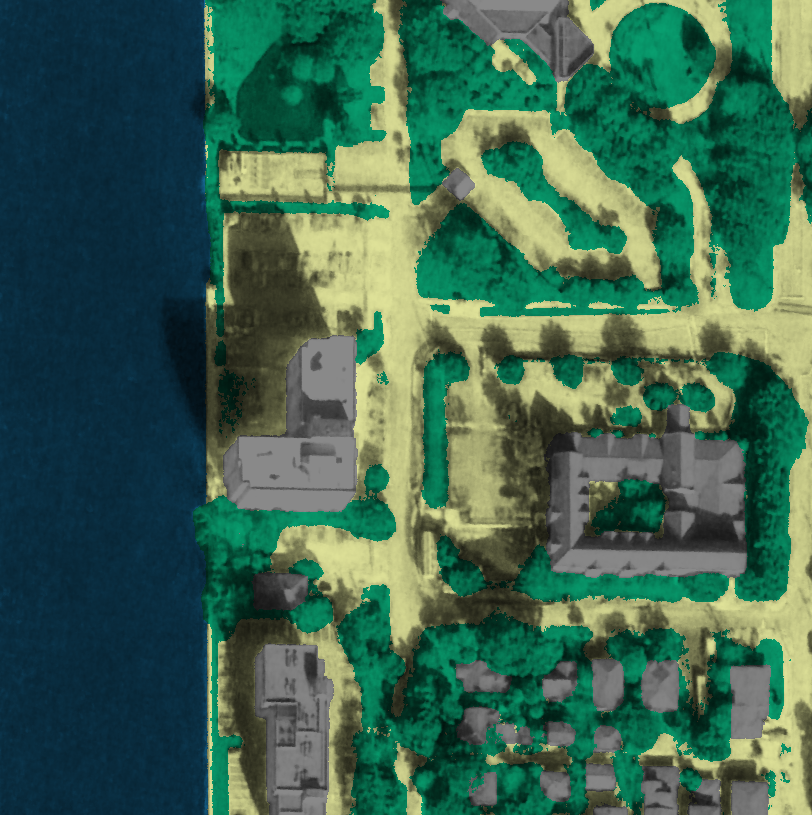}}%
	\end{minipage}\\

	\caption{
		Qualitative results of our proposed \emph{NeRF} model for the popular scenes of the JAX dataset. The pixel-wise semantic annotations in (b) are part of our dataset.
		The domain-adapted \emph{NeRF} model is able to learn and reproduce 
		a representation of the scene containing color and semantic information. By combining the predicted semantic class with the learned lighting component we enrich the visualization with three-dimensional depth cues. }
	\label{fig:results}
\end{figure*}

The publicly available \textit{Urban Semantic 3D} \cite{us3d} dataset provides automatically generated labels for the \emph{Data-Fusion-Contest 2019 Track-3} (\emph{DFC2019}) \cite{dfc2019} dataset. 
The generation process fuses information from LiDAR source data with limited supervision, leading to a high degree of noise and poor accuracy.
Transient objects such as \emph{Vehicles} are not captured in the provided labels.
We therefore chose to produce our own annotations and ensure high quality annotations through a high degree of manual refinement. 
We make it publicly available to further advance research in the field of semantic 3D reconstruction.

Our semantic multi-view satellite dataset contains manually generated annotations for 71 satellite images across four scenes. 
Analog to previous work \cite{snerf,satnerf,eonerf} we select a popular subset containing four scenes ($004, 068,214,260$) of the Jacksonville (Florida, USA) area. The provided Maxar WorldView-3 satellite images are captured between $2014$ and $2016$ with a $0.3$m/pixel resolution at nadir.

The annotations are generated as pixel-wise segmentation maps and cover the main $256 \times 256$m area-of-interest of each scene. This area is defined by the LiDAR \emph{Digital Surface Model} (\emph{DSM}) contained in  the \emph{DFC2019} \cite{dfc2019} dataset. 
To create the annotations we manually refine initial rough estimates of a user guided, class agnostic foundation model for semantic segmentation \cite{sam}. We manually verify and potentially correct all annotations to ensure high accuracy and consistency across all images. 
The five following classes were selected based on scene content:
\emph{Ground}, \emph{Water}, \emph{Vegetation}, \emph{Buildings} and \emph{Vehicles}.
We show one satellite image per scene in \cref{fig:results_rgb} and its accompanying pixel-wise segmentation in \cref{fig:results_labels}.

\section{Implementation and Training Details}

Analog to \emph{SatNeRF} \cite{satnerf} we use $8$ layers of $512$ features for the main feature backbone of the MLP. The additional semantic head consists of a single hidden layer with $256$ features and uses a $\mathit{sigmoid}$ activation function for semantic logit prediction. The architecture of the other heads such as the lighting scalar $\mathit{sun}$ or the uncertainty $\beta$ are identical to \emph{SatNeRF} \cite{satnerf}.

The semantic loss weight is set to $\lambda_s = 0.04$ as suggested by \emph{Semantic-NeRF} \cite{semanticnerf}.
A transient regularization loss value of $\lambda_t = 0.1$ shows in our experiments a good compromise between guiding the network to filter transient objects while simultaneously applying the uncertainty $\beta$ for other color inconsistencies.
The transient embedding $\vec{t}_j$ has a dimension of $4$.
As proposed by \emph{SatNeRF} \cite{satnerf} we use additional depth supervision rays for the first $25\%$ of training iterations with a weight of $\lambda_{DS} = 1000$ and solar correction rays with a weight of $\lambda_{SC} = 0.05$.
Depth supervision refers to using a sparse pointcloud created during the inital \emph{Bundle Adjustment} on the \emph{RPC} camera models for depth guidance. 
Solar correction rays are used as part of the satellite-domain-adapted lighting model to align shadows with the scene structure based on the density.

We trained all models in this work for $300k$ iterations with an Adam optimizer starting with a learning rate of $5e^{-4}$, which is decreased by a factor of $0.9$ every epoch. Each ray is discretized into $64$ points and we process $1024$ rays each training iteration. 
All models were trained using a NVIDIA RTX 4090 with $24$ GB of RAM.

\section{Evaluation and Experiments}

\begin{table}[t]
	\centering
	\setlength{\tabcolsep}{0.5em} %
	\begin{tabular}{l|cccc|c}
		\toprule
		& \multicolumn{5}{c}{\textbf{Semantic Accuracy $\uparrow$}} \\
		Split &004&068&214&260& Mean\\
		\midrule
		
		Train & 0.936 & 0.950 & 0.947 & 0.934& 0.942 \\
		Test & 0.930 & 0.948 & 0.944 & 0.916& 0.935\\
		
		\bottomrule
	\end{tabular}
	\caption{Quantitative evaluation of our domain-adapted semantic \emph{NeRF}. The high semantic accuracy for test views shows the ability of the network to render a correct semantic segmentation of the scene for novel, during training unseen views. }
	\label{tbl:semantic_results}
\end{table}

We evaluate the ability of our proposed \emph{NeRF} pipeline to learn a generalized semantic understanding of satellite scenes. 
Since the \emph{Vehicle} category consists of transient objects, we treat this class according to \cref{subsec:handling_transients} when computing the semantic and color representation

In \cref{subsec:learned_semantic_rep} we evaluate the general quality of the semantic rendering in comparison to ground truth data. In \cref{subsec:transient_reg} we show the impact of our transient regularization term for color representation and how it reduces artifacts. 
Lastly, we evaluate the capability of our pipeline to improve the semantic segmentation quality of input images by leveraging the inherent multi-view-consistency provided by \emph{NeRF} in \cref{subsec:multi_view_consistency}.

\subsection{Learned Semantic Representation}
\label{subsec:learned_semantic_rep}

To evaluate the quality of the learned semantic representation, we use the semantic accuracy metric, \ie the ratio of correctly to incorrectly identified pixels. 
We create alternative transient-free ground truth annotations to assess whether our proposed network is able to accurately replace the missing semantic information for transient locations with the correct static class (in most cases either \emph{Ground} or \emph{Building}) . 
The evaluation results are reported in \cref{tbl:semantic_results}.
Our domain-adapted semantic \emph{NeRF} achieves an accuracy of over 90\% for all scenes, proving its ability to generalize the learned semantic understanding of the scene for both training and testing viewpoints.

We visualize one view of each scene in \cref{fig:results} 
with the corresponding ground truth color image (\cref{fig:results_rgb}), the semantic annotations (\cref{fig:results_labels}), and the outputs learned by our domain-adapted semantic \emph{NeRF}.
Comparing the learned semantic prediction in \cref{fig:results_semantic} with the ground truth annotations shows a high degree of accuracy. 
By combining the predicted semantic class with the learned shadow scalar in \cref{fig:results_shadows} we produce a three-dimensional visualization in \cref{fig:results_semantic_shaded}.
Comparing this visualization to the ground truth color image shows how the network is able to transfer  the scene structure on the semantic understanding.

\begin{table}[t]
	\centering
	\setlength{\tabcolsep}{0.5em} %
	\begin{tabular}{c|cccc|c}
		\toprule
		\textbf{Transient} & \multicolumn{5}{c}{\textbf{Transient Uncertainty $\uparrow$}}\\
		\textbf{Regularization} &004&068&214&260&Mean\\
		\midrule
		
		\crossmark & 0.121 & 0.092 & 0.144 & 0.027 & 0.096\\
		
		\checkmark & 0.976 & 0.900 & 0.877 & 0.915 & 0.917\\
		\bottomrule
	\end{tabular}
	\caption{
		Quantitative evaluation of the proposed transient regularization loss. The results shows a great increase in uncertainty for locations marked as transient in the semantic ground truth data. 
		The model learns a static, transient free appearance of the scene. }
	\label{tbl:semantic_results_car_filtering}
\end{table}

\begin{figure*}[ht!]
	\centering
	
	\begin{minipage}{0.025\linewidth}
		\centering
		\rotatebox[origin=center]{90}{SatNeRF \cite{satnerf}}
	\end{minipage}
	\begin{minipage}{0.97\linewidth}
		\centering
		\includegraphics[width=.24\linewidth]{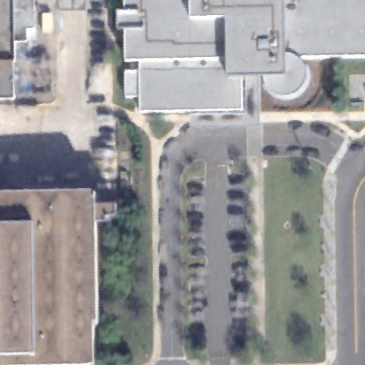}
		\hfill
		\includegraphics[width=.24\linewidth]{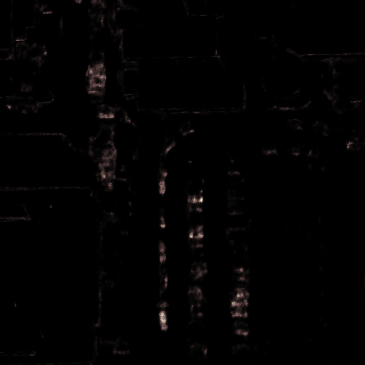}
		\hfill
		\includegraphics[width=.24\linewidth]{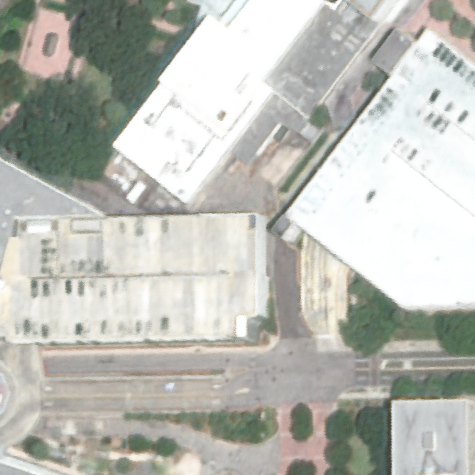}
		\hfill
		\includegraphics[width=.24\linewidth]{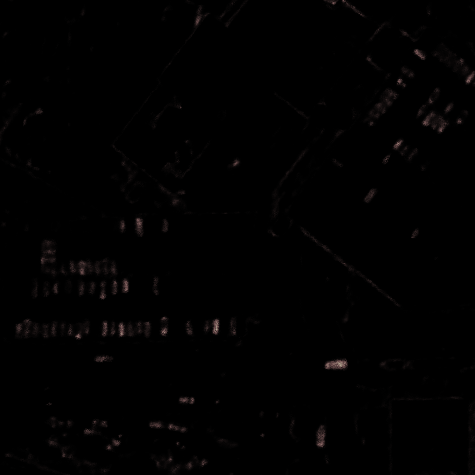}
	\end{minipage}\\
	
	\vspace{0.2 cm}
	\begin{minipage}{0.02\linewidth}
		\centering
		\rotatebox[origin=center]{90}{Ours}
	\end{minipage}
	\begin{minipage}{0.975\linewidth}
		\centering
		\subcaptionbox{JAX\_068}{
			\includegraphics[width=.24\linewidth]{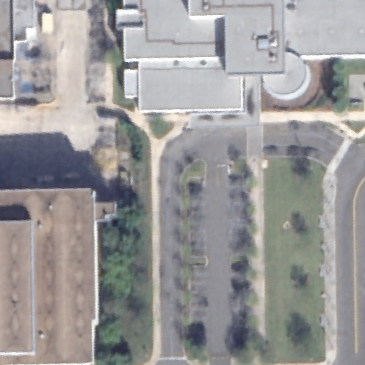}
			\hfill
			\includegraphics[width=.24\linewidth]{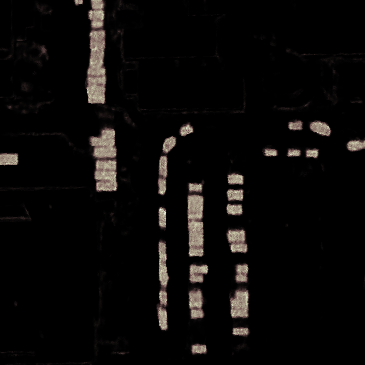}
		}%
		\hfill
		\hspace{0.02cm}
		\subcaptionbox{JAX\_214}{
			\includegraphics[width=.24\linewidth]{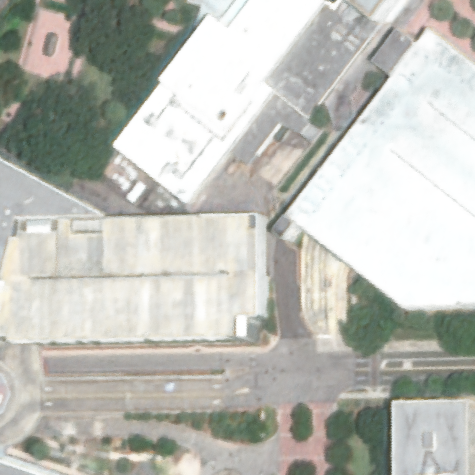}
			\hfill
			\includegraphics[width=.24\linewidth]{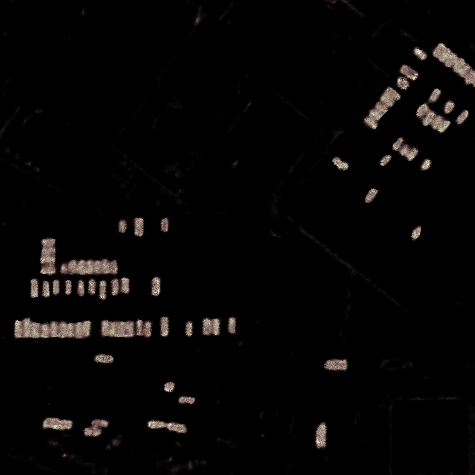}
		}%
		
	\end{minipage}\\

	\caption{
		Qualitative comparison of the proposed transient regularization loss. Using this loss, our model is able to substantially reduce blurry artifacts by guiding the uncertainty $\beta$ on locations of transient objects.
		Shown here are closeups of the rendered color image and visualization of the uncertainty scalar $\beta$ for training views of the scenes JAX\_068 and JAX\_214. The Baseline \emph{SatNeRF} \cite{satnerf} model is unable to filter the cars out, leading to visible remnants. Our method is able to render a static, transient free representation of the scene.}
	\label{fig:car_reg_loss}
\end{figure*}

\subsection{Transient Regularization}
\label{subsec:transient_reg}

We evaluate the impact of our proposed transient regularization loss in increasing uncertainty of known transient locations by measuring the \emph{Transient Uncertainty}.
We define this metric as the mean uncertainty for all transient locations $\mathcal{R}^t$, \ie the \emph{Vehicles} class, in the semantic ground truth data. The uncertainty $\beta(\vec{x}, \vec{t}_j)$ is aggregated along rays $\vec{r}$ analog to \cref{eq:semantic_rendering} with the per-image embedding $\vec{t}_j$ as additional input. 
\begin{equation}
	\beta_t(\mathcal{R}^t) = \frac{1}{|\mathcal{R}^t|}\sum_{\vec{r} \in \mathcal{R}^t} \sum_{i = 0}^{N} \alpha_i T_i \beta(\vec{x}_i, \vec{t}_j)
\end{equation}
Since the uncertainty $\beta$ is only relevant during training, we do not consider test images during this evaluation. 
In \cref{tbl:semantic_results_car_filtering} we can see the impact of our transient regularization loss. 
The baseline \emph{SatNeRF} \cite{satnerf} model is not able to adequately capture the transient objects, resulting in a low \emph{Transient Uncertainty} of $0.096$ across all four scenes. 
This causes the \emph{SatNeRF} model to integrate the corresponding appearance information into the color representation.
Our proposed transient regularization loss greatly increases the \emph{Transient Uncertainty} to $0.917$, showing the benefit from the additional guidance. 

We provide close up visualizations of the uncertainty $\beta$ and the corresponding rendered color image for two scenes in \cref{fig:car_reg_loss}. Both contain parking lots with large numbers of vehicles parked across multiple training views.
For both scenes the uncertainty $\beta$ shows increased focus using our transient regularization loss, resulting in a drastic decrease in visible vehicles in the rendered color image. 
Only the static elements of the scene are rendered by the model.

\subsection{Multi-View Consistency}
\label{subsec:multi_view_consistency}

\begin{table*}[t]
	\centering
	\begin{tabular}{l|c|cccc|c}
		\toprule
		& & \multicolumn{4}{c}{\textbf{Semantic Accuracy (Train) $\uparrow$}} &  \\
		Labels & Activation Function & JAX\_004&JAX\_068&JAX\_214&JAX\_260& mean\\
		\midrule
		
		Corrupted & - & 0.800 & 0.800 & 0.800 & 0.800 & 0.800\\
		
		Fused Labels & None & 0.857 &  0.930 & 0.924 & 0.902 & 0.903\\
		
		Fused Labels & Sigmoid & 0.894 &  0.936 & 0.933 & 0.920 & 0.921\\
		
		\bottomrule
	\end{tabular}
	\caption{
		Quantitative evaluation of the multi-view consistency on imperfect labels.
		Enforced multi-view consistency is able to improve initial semantic segmentations used as training input data. By using a sigmoid activation function for semantic logit prediction we can increase the semantic accuracy by a mean of $12.1\%$. }
	\label{tbl:corrupted_labels}
\end{table*}

\begin{figure*}[t]
	\centering
	
	\begin{minipage}{0.035\linewidth}
		\centering
		\rotatebox[origin=center]{90}{JAX\_068}
	\end{minipage}
	\begin{minipage}{0.95\linewidth}
		\centering
		\hspace{-0.13cm}
		\includegraphics[width=.24\linewidth]{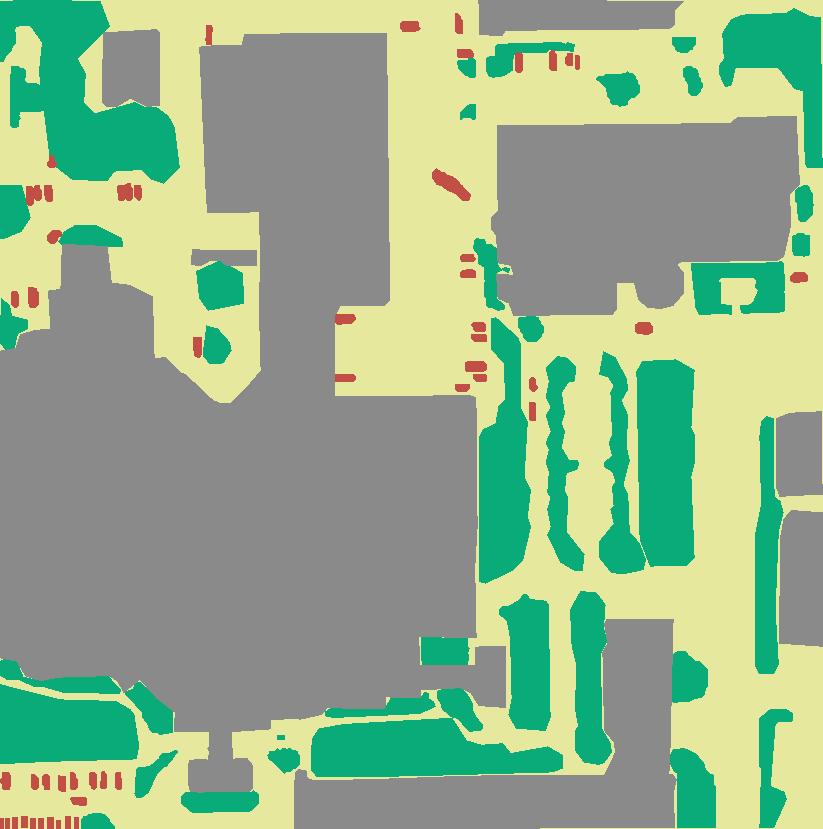}
		\hfil
		\includegraphics[width=.24\linewidth]{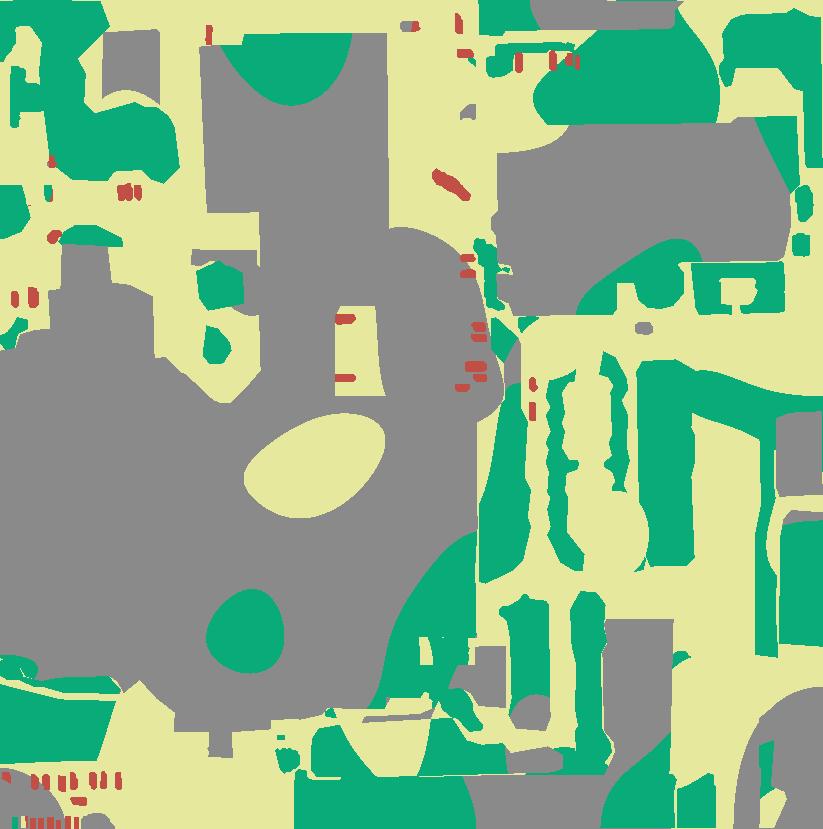}
		\hfil
		\includegraphics[width=.24\linewidth]{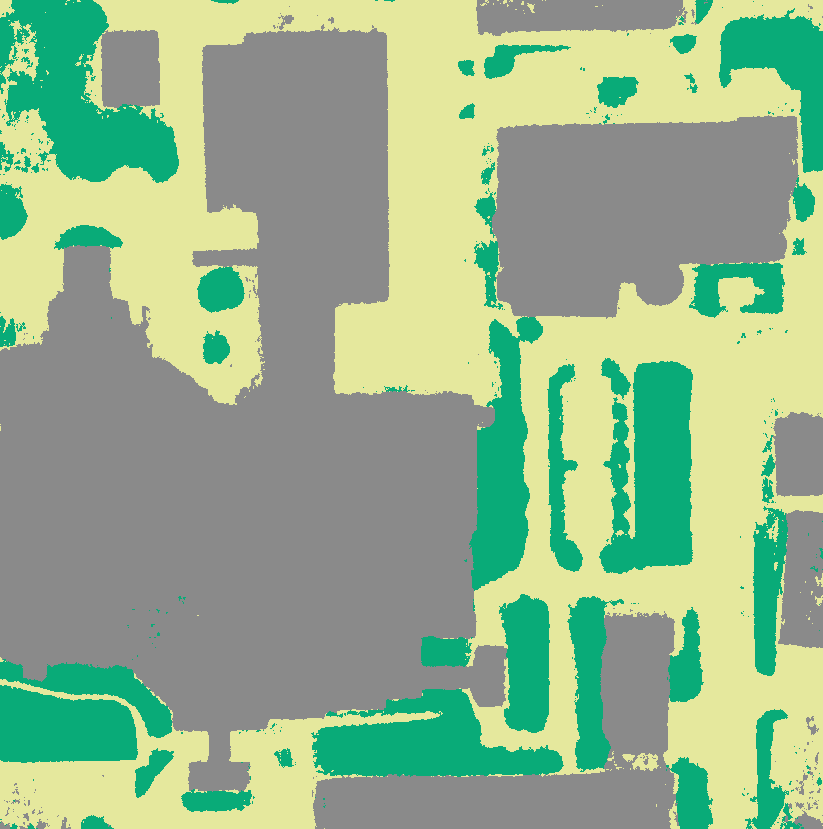}
		\hfil
		\includegraphics[width=.24\linewidth]{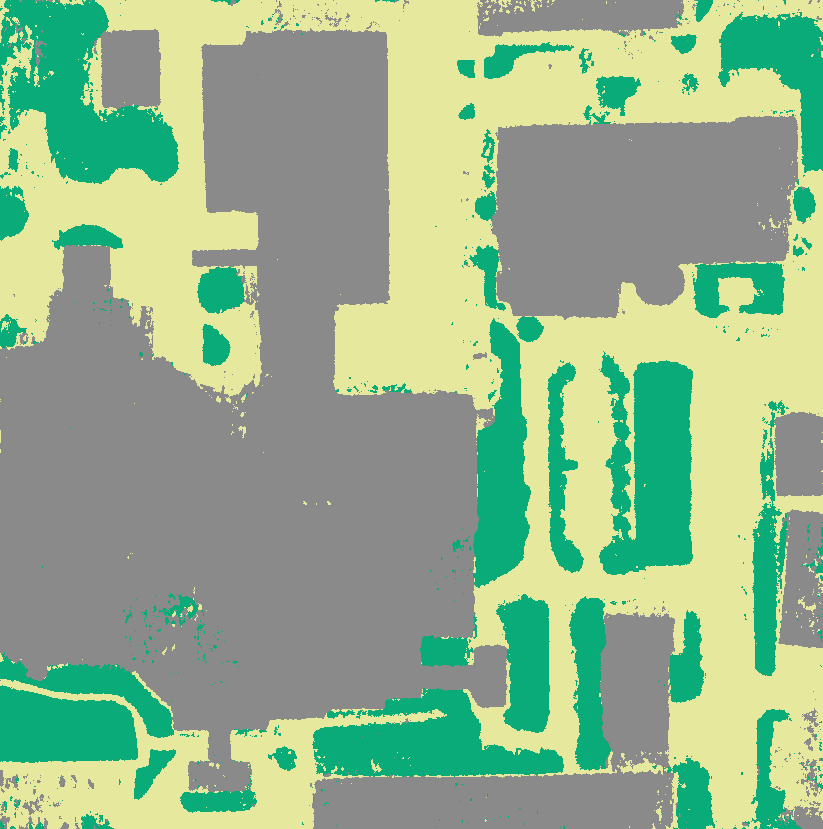}
	\end{minipage}
	\\
	\vspace{0.2cm}
	\begin{minipage}{0.04\linewidth}
		\centering
		\rotatebox[origin=center]{90}{JAX\_214}
	\end{minipage}
	\begin{minipage}{0.95\linewidth}
		\begin{subfigure}{0.24\textwidth}
			\centering
			\includegraphics[width=\textwidth]{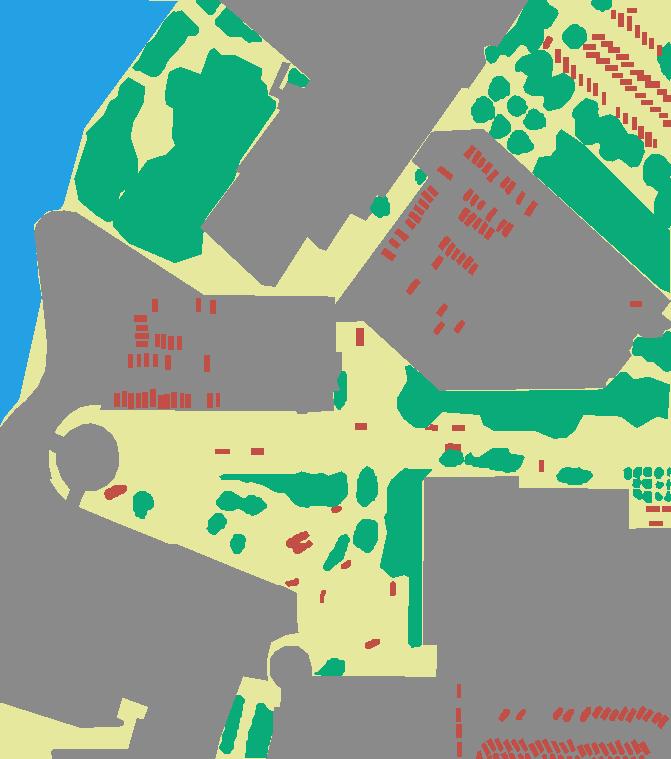}
			\caption{Ground Truth.}
			\label{subfig:lbl_corruption_gt}
		\end{subfigure}
		\hfil
		\begin{subfigure}{0.24\textwidth}
			\centering
			\includegraphics[width=\textwidth]{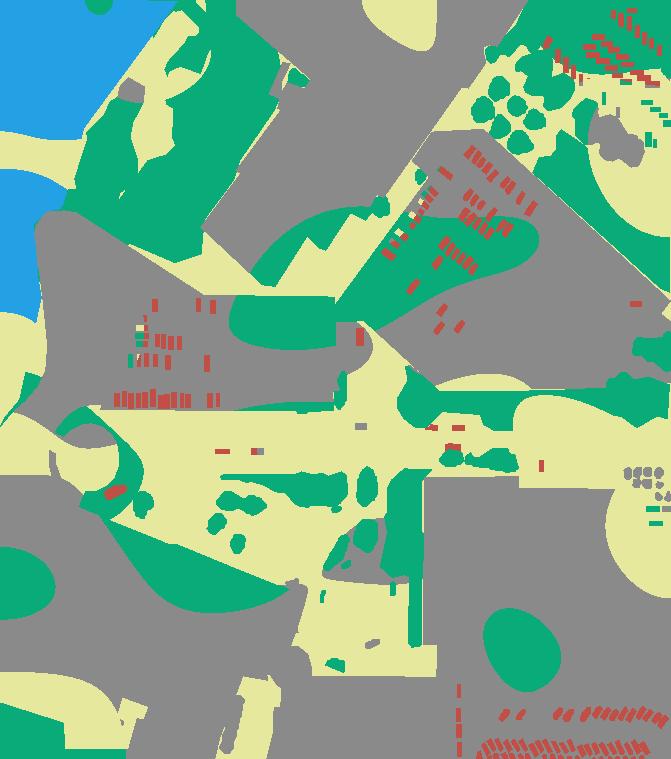}
			\caption{Simulated Segmentation.}
			\label{subfig:lbl_corruption_corrupted}
		\end{subfigure}
		\hfil
		\begin{subfigure}{0.24\textwidth}
			\centering
			\includegraphics[width=\textwidth]{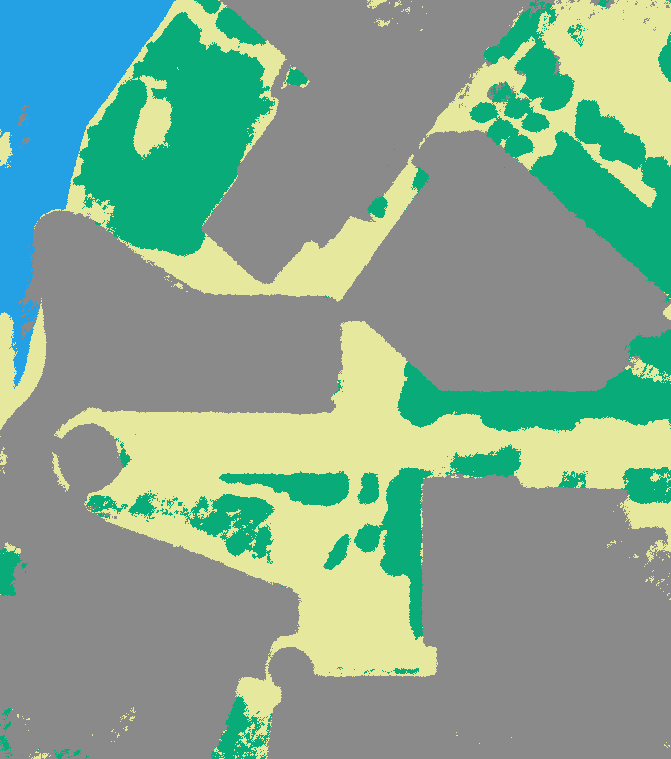}
			\caption{Fused Labels with Sigmoid.}
			\label{subfig:lbl_corruption_nerf_output}
		\end{subfigure}
		\hfil
		\begin{subfigure}{0.24\textwidth}
			\centering
			\includegraphics[width=\textwidth]{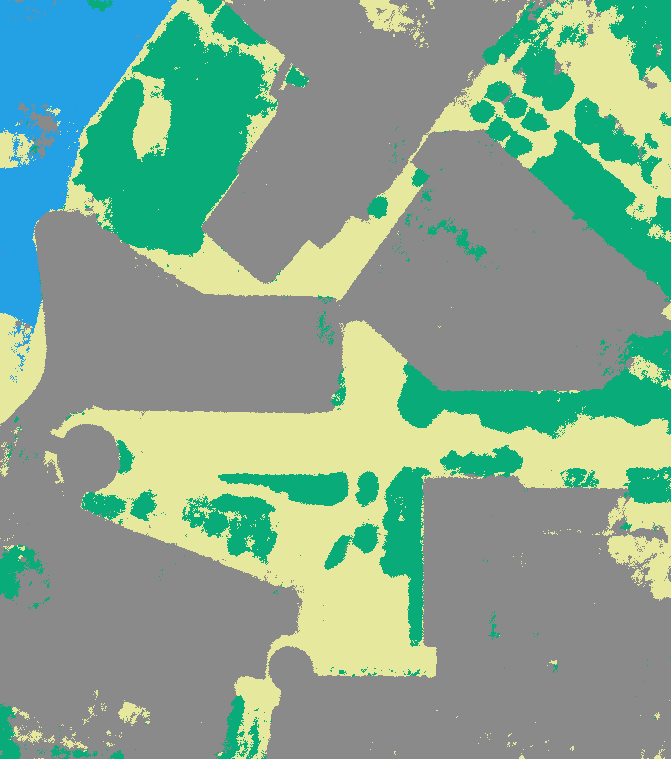}
			\caption{Fused Labels w/o Sigmoid.}
			\label{subfig:lbl_corruption_nerf_output_no_activation}
		\end{subfigure}
	\end{minipage}
	\caption{Effect of the multi-view consistency on local segmentation errors.  
		The semantic information across all training views is merged in order to improve accuracy.
		Using a $\mathit{sigmoid}$ activation function as normalization 	decreases noise in the reconstructed segmentation. }
	\label{fig:s_nerf_shading}
\end{figure*}

\emph{NeRF} learn a generalized scene understanding through maximizing consistency across all training views. 
This multi-view consistency reflects an averaging of information between input images. 
The semantic prediction potentially benefits from this process, improving semantic segmentation quality through enforcing multi-view consistency.

\emph{Semantic NeRF} \cite{semanticnerf} conducts multiple experiments regarding the effect of partial or corrupted labels. 
They show that the network is able to learn a meaningful semantic representation, even if pixel-wise noise is added or entire instances are flipped to a different class. 
With these experiments they demonstrate that the underlying multi-view consistency enforced by \emph{NeRF} enables the generation of coherent semantic renderings even when training with partial and/or faulty labels. 
The \emph{NeRF} allows to denoise and improve the initial semantic segmentations by rendering the same viewpoint. 
We conduct additional experiments researching this use case for the domain of segmenting satellite data.

We degrade the quality of our training data to more closely match the predicted output of automatic segmentation methods. 
We aim to simulate the effect of automatic segmentation methods correctly recognizing and segmenting most of an object, with smaller areas being incorrectly classified. 
We therefore explicitly refrain from pixel-wise noise and instance specific manipulation in contrast to \emph{Semantic-NeRF} \cite{semanticnerf}.
To corrupt our labels we generate a normal noise map for each semantic class. 
We blur and rescale the noise maps to create larger regions.  
Based on a specified accuracy loss, thresholds are determined to identify regions in each noise map that should be corrupted. 
Each noise region is reassigned to a different class.  
We release the corrupted labels alongside the code for the automated label corruption as part of the dataset.
To evaluate the impact of training our pipeline with corrupted labels, we degrade the annotations for each scene by $20\%$, resulting in an overall semantic accuracy of $80\%$. 
The impact on the pixel-wise labels is illustrated in \cref{subfig:lbl_corruption_gt,subfig:lbl_corruption_corrupted}.

Rendering the same training views with our fully trained \emph{NeRF} model leads to a considerable improvement in semantic accuracy. The results are reported in \cref{tbl:corrupted_labels}. 
Across all four scenes the semantic accuracy increases by a mean of $12.1\%$ when using a sigmoid activation function for the semantic logits $\vec{s}$. If using no activation function, as originally proposed by \emph{Semantic-NeRF} \cite{semanticnerf}, the semantic accuracy increases only by $10.3\%$.
Comparing the predicted semantic output of the \emph{NeRF} in \cref{subfig:lbl_corruption_nerf_output} with the corrupted labels in \cref{subfig:lbl_corruption_corrupted} shows the ability of the model to largely replace and fix the holes caused by the automatic label degradation. Using no activation function leads to a higher amount of noise as visible in \cref{subfig:lbl_corruption_nerf_output_no_activation}.

\section{Conclusions}

This paper proposes a novel \emph{NeRF} architecture for generating a unified color and semantic 3D representation of scenes derived from satellite images and their corresponding pixel-wise labels. 
By leveraging semantic information of transient categories during the training process we are able to improve the appearance representation of corresponding areas.
This allows the NeRF to learn a more generalized and artifact-free representation of the scene. 
Our method allows to enhance the quality of noisy input labels by exploiting the multi-view consistency offered by modern \emph{NeRF} models.
To support future research in this area, we provide a dataset featuring manually generated labels for four multi-view scenes. 
Additionally, we release the code for this work, including scripts for model training and evaluation.

{\small
\bibliographystyle{ieee_fullname}
\bibliography{Semantic_RS_NeRF}

\begin{thebibliography}{10}\itemsep=-1pt

\bibitem{sundial}
Nikhil Behari, Akshat Dave, Kushagra Tiwary, William Yang, and Ramesh Raskar.
\newblock Sundial: 3d satellite understanding through direct, ambient, and
  complex lighting decomposition, 2023.

\bibitem{ames}
Ross~A. Beyer, Oleg Alexandrov, and Scott McMichael.
\newblock The ames stereo pipeline: Nasa's open source software for deriving
  and processing terrain data.
\newblock {\em Earth and Space Science}, 5(9):537--548, 2018.

\bibitem{us3d}
Marc Bosch, Kevin Foster, Gordon Christie, Sean Wang, Gregory~D Hager, and
  Myron Brown.
\newblock Semantic stereo for incidental satellite images, 2018.

\bibitem{ssr}
Sebastian Bullinger, Christoph Bodensteiner, and Michael Arens.
\newblock 3d surface reconstruction from multi-date satellite images, 2021.

\bibitem{s2p}
C. de Franchis, E. Meinhardt-Llopis, J. Michel, J.-M. Morel, and G. Facciolo.
\newblock An automatic and modular stereo pipeline for pushbroom images.
\newblock {\em ISPRS Annals of the Photogrammetry, Remote Sensing and Spatial
  Information Sciences}, II-3:49--56, 2014.

\bibitem{snerf}
Dawa Derksen and Dario Izzo.
\newblock Shadow neural radiance fields for multi-view satellite
  photogrammetry.
\newblock {\em CoRR}, abs/2104.09877, 2021.

\bibitem{pinhole}
R.~I. Hartley and A. Zisserman.
\newblock {\em Multiple View Geometry in Computer Vision}, chapter 6: Camera
  Models.
\newblock Cambridge University Press, ISBN: 0521540518, second edition, 2004.

\bibitem{sam}
Alexander Kirillov, Eric Mintun, Nikhila Ravi, Hanzi Mao, Chloe Rolland, Laura
  Gustafson, Tete Xiao, Spencer Whitehead, Alexander~C. Berg, Wan-Yen Lo, Piotr
  Doll{\'a}r, and Ross Girshick.
\newblock Segment anything.
\newblock {\em arXiv:2304.02643}, 2023.

\bibitem{dfc2019}
Bertrand Le~Saux, Naoto Yokoya, Ronny Hänsch, and Myron Brown.
\newblock Data fusion contest 2019 (dfc2019), 2019.

\bibitem{eonerf}
Roger Mar{\'\i}, Gabriele Facciolo, and Thibaud Ehret.
\newblock Multi-date earth observation nerf: The detail is in the shadows.
\newblock In {\em Proceedings of the IEEE/CVF Conference on Computer Vision and
  Pattern Recognition (CVPR) Workshops}, pages 2034--2044, June 2023.

\bibitem{satnerf}
Roger Marí, Gabriele Facciolo, and Thibaud Ehret.
\newblock Sat-nerf: Learning multi-view satellite photogrammetry with transient
  objects and shadow modeling using rpc cameras, 2022.

\bibitem{nerf}
Ben Mildenhall, Pratul~P. Srinivasan, Matthew Tancik, Jonathan~T. Barron, Ravi
  Ramamoorthi, and Ren Ng.
\newblock Nerf: Representing scenes as neural radiance fields for view
  synthesis, 2020.

\bibitem{pointnet++}
Charles~R Qi, Li Yi, Hao Su, and Leonidas~J Guibas.
\newblock Pointnet++: Deep hierarchical feature learning on point sets in a
  metric space.
\newblock {\em arXiv preprint arXiv:1706.02413}, 2017.

\bibitem{rpc}
C.~Vincent Tao and Yong Hu.
\newblock A comprehensive study of the rational function model for
  photogrammetric processing.
\newblock {\em Photogrammetric Engineering and Remote Sensing}, 67:1347--1357,
  2001.

\bibitem{kpconv}
Hugues Thomas, Charles~R. Qi, Jean-Emmanuel Deschaud, Beatriz Marcotegui,
  François Goulette, and Leonidas~J. Guibas.
\newblock Kpconv: Flexible and deformable convolution for point clouds, 2019.

\bibitem{pt3}
Xiaoyang Wu, Li Jiang, Peng-Shuai Wang, Zhijian Liu, Xihui Liu, Yu Qiao, Wanli
  Ouyang, Tong He, and Hengshuang Zhao.
\newblock Point transformer v3: Simpler, faster, stronger, 2024.

\bibitem{vissat}
Kai Zhang, Jin Sun, and Noah Snavely.
\newblock Leveraging vision reconstruction pipelines for satellite imagery,
  2019.

\bibitem{beyondrgb}
Mingtong Zhang, Shuhong Zheng, Zhipeng Bao, Martial Hebert, and Yu-Xiong Wang.
\newblock Beyond rgb: Scene-property synthesis with neural radiance fields,
  2022.

\bibitem{semanticnerf}
Shuaifeng Zhi, Tristan Laidlow, Stefan Leutenegger, and Andrew~J. Davison.
\newblock In-place scene labelling and understanding with implicit scene
  representation.
\newblock {\em CoRR}, abs/2103.15875, 2021.

\end{thebibliography}
}

\end{document}